\documentclass[%
reprint,
amsmath,amssymb,
pra,
]{revtex4-2}

\usepackage{graphicx}% Include figure files
\usepackage{dcolumn}% Align table columns on decimal point
\usepackage{bm}% bold math
\usepackage{cleveref}
\newcommand{\real}{\mathbb{R}}
\newcommand{\gauss}{\mathcal{N}}

\newcommand{\reward}{r}
\newcommand{\policy}{\pi}
\newcommand{\mdp}{\mathcal{M}}
\newcommand{\states}{\mathcal{S}}
\newcommand{\actions}{\mathcal{A}}

\newcommand{\transitions}{T}
\newcommand{\initstate}{d_0}

\newcommand{\horizon}{H}

\newcommand{\discount}{\gamma}

\newcommand{\bellman}{\mathcal{B}}

\newcommand{\traj}{\tau}

\newcommand{\bs}{\mathbf{s}}
\newcommand{\ba}{\mathbf{a}}

\usepackage{tikz}
\usetikzlibrary{arrows,positioning} 
\tikzset{
	>=stealth',
	punkt/.style={
		rectangle,
		rounded corners,
		draw=black, very thick,
		text width=3.5em,
		minimum height=2em,
		text centered},
	pil/.style={
		->,
		thick,
		shorten <=2pt,
		shorten >=2pt,}
}
\begin{document}
	
	\preprint{APS/123-QED}
	
	\title{Model-free and Bayesian Ensembling Model-based Deep Reinforcement Learning for Particle Accelerator Control Demonstrated on the FERMI FEL}% Force line breaks with \\
	
	\author{Simon Hirlaender}
	\affiliation{Paris Lodron University, Salzburg, AT}
	\email{simon.hirlaender@plus.ac.at}%Lines break automatically or can be forced with \\
	%	\author{Niky Bruchon}%
	%	\email{Second.Author@institution.edu}
	%	\affiliation{%
		%		Authors' institution and/or address\\
		%		This line break forced with \textbackslash\textbackslash
		%	}%
	%	
	%	\collaboration{MUSO Collaboration}%\noaffiliation
	
	\author{Niky Bruchon}
	\affiliation{CERN, Geneva, CH}% with \\
	%
	%	\author{Delta Author}
	%	\affiliation{%
		%		Authors' institution and/or address\\
		%		This line break forced with \textbackslash\textbackslash
		%	}%
	%	
	%	\collaboration{CLEO Collaboration}%\noaffiliation
	
	\date{\today}% It is always \today, today,
	
	\begin{abstract}
		Reinforcement learning holds tremendous promise in accelerator controls. The primary goal of this paper is to show how this approach can be utilised on an operational level on accelerator physics problems. Despite the success of model-free reinforcement learning in several domains, sample-efficiency still is a bottleneck, which model-based methods might encompass.
		We compare well-suited purely model-based to model-free reinforcement learning applied to the intensity optimisation on the FERMI FEL system. We find that the model-based approach demonstrates higher representational power and sample-efficiency, while the asymptotic performance of the model-free method is slightly superior. The model-based algorithm is implemented in a DYNA-style using an uncertainty aware model, and the model-free algorithm is based on tailored deep Q-learning. The algorithms were designed to present increased noise robustness as omnipresent in accelerator control problems in both cases. Code is released in \href{https://github.com/MathPhysSim/FERMI_RL_Paper}{https://github.com/MathPhysSim/FERMI\_RL\_Paper}. 
		%\begin{description}
		%\end{description}
	\end{abstract}
	
	\maketitle
	
	%\tableofcontents
	
	\section{Introduction}\label{sec:introduction}
	
\begin{figure}[ht]
		\centering
		\includegraphics*[width=0.4\textwidth]{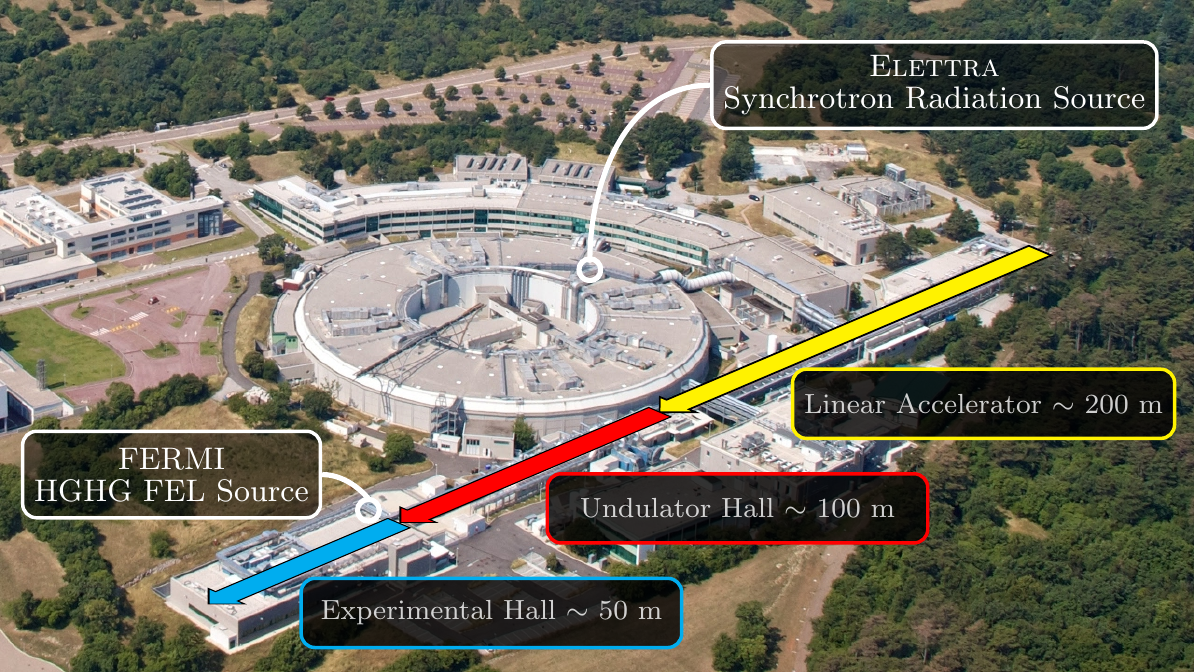}
		\caption{The Elettra research centre hosting the FERMI free electron laser \cite{Brochon2020}.}
		\label{fig:elletra_research}
	\end{figure}

	In particle accelerator operation, one main goal is to provide stable and reproducible performance. Achieving this demands the consideration of several control problems simultaneously \cite{Kain2020,Scheinker2018}. Especially if there is no way to model the physics a priori, one might use optimisation techniques as, e.g. derivative-free optimisers (DFOs) \cite{Huang2013,Bruchon2017,Scheinker2020,Hirlaender2019,Welsch2015,Albright2019} or model-based optimisations as Gaussian processes \cite{Hanuka2020,Roussel2020} to restore or maintain the performance.
	%\\
	%
	Another way to face this challenging task is proposed in \cite{Bruchon2021}, where, within the control theory field, an online iterative Linear Quadratic Regulator (iLQR) is applied in a Model-Predictive Control fashion using a neural network trained on real data to perform a model identification.
	\\
	Recently, the community has begun to explore a novel approach to solve control problems. Reinforcement learning (RL) \cite{Bruchon2020,Bruchon2019,Kain2020,Pang2020,John2020,O2020} unveils several advantages over optimisation methods:
	\begin{itemize}
		\item It covers a larger class of problems, as RL optimises a sequence of decisions.
		\item It memorises the problem and does not always begin from zero as a DFO.
		\item Already existing data can be used.
		%\item Adaption to non-stationary systems is possible.
		\item The underlying structure of the problem might be deduced.
	\end{itemize}
	One demanding aspect using RL in real-world applications is the number of iterations needed to train a controller - the sample-efficiency, since RL methods are known to be data-hungry \cite{Sutton2018,DulacArnold2019}. A second critical aspect to be considered is the robustness of the training in a noisy data-limited environment.\\
	In this paper, we present the study carried out to solve the maximisation-problem of the radiation intensity generated by a seeded Free-Electron Laser (FEL) on the Free Electron laser Radiation for Multidisciplinary Investigations (FERMI) at \emph{Elettra Sincrotrone Trieste}. Fig.~\ref{fig:elletra_research} shows an aerial
	overview of the Elettra site and the FERMI free-electron laser buildings.\\
	We successfully applied two different algorithm classes, a highly sample-efficient variant of continuous Q-learning (Section~\ref{ss:Normalized advantage function}) and a newly proposed model-based algorithm, which we call \emph{AE-DYNA} (Section~\ref{ss:Uncertainty aware DYNA-style reinforcement learning}), to the FERMI FEL problem.\\ The methods reveal different characteristics addressing the stated critical aspects. They are generally applicable to problems showing a sample-complexity as in common accelerator control problems.
	\subsection{An Overview of the Main Results}
	We demonstrate optimisation techniques that simultaneously adjust four parameters, the tilt and incline of two mirrors that control the trajectory and overlap of the seed laser in the FERMI FEL relative to the beam in the undulator. 
	%	We show that our algorithms require significantly less than 1000 training sets, after which they can maximise the FERMI FEL's output intensity to $>$ that 95\% of the maximum within only $\sim 3$ to $\sim 5$ steps depending on which algorithm is used.
	By making use of model ensembles and anchored neural networks our algorithms exhibit a high sample-efficiency (learning using $<1000$ data sets), a low model-bias, noise resistivity and are extremely fast as they are able to maximize the FERMI FEL's output intensity to greater than 95\% of the maximum within only 3-5 steps, whereas the maximal intensity was determined through manual expert optimisation.
	Thus, we show the operational applicability of purely model-based reinforcement learning in an accelerator control problem.\\
	Furthermore, our algorithms can perform real-time learning based on data and therefore, can be re-run and re-trained for time-varying systems.\\
	The paper is organised as follows:
	\begin{itemize}
		\item Description of the problem set-up at FERMI
		\item Overview of RL
		\item Design decisions of the implementations used in these studies and theoretical concerns
		\item Results of the experiments 
		\item Discussion, outlook and summary
	\end{itemize}
	
	\section{The set-up of the studied problem}
	\subsection{The Physical Set-up}
	In a seeded free-electron laser one of the most critical parameters is the temporal and transverse overlap of the electron and laser beam in the magnetic section called modulator undulator.\\
	%	At FERMI, several beam-based feedback systems are deployed to control the beams trajectories shot to shot with the primary purpose of guaranteeing a steady intensity of the light radiation. 
	In the last years, various approaches have been studied to investigate their applicability in machine tuning. \\
	A free-electron laser is a fourth-generation light source where the lasing medium consists of a very-high-speed electron moving freely through a magnetic structure. By using an external seeding source, the FERMI FEL has several advantages relative to standard FEL approaches in terms of increased stability in pulse and photon energy, reduced size, improved longitudinal coherence, more control on the pulse structure, and improved temporal synchronisation with external timing systems.\\
	The external optical laser signal interacts with the relativistic electron beam in the undulator, introducing an energy modulation that aids in the FEL process. The modulation in energy is converted into a charge modulation in the dispersive section, and finally, the density modulated beams radiation is amplified in the section of the radiator.\\
	The importance of ensuring the best possible overlap between the seed laser and the electron beam in the modulator is therefore evident.\\
	%	For this reason, the proposed study focuses on the control of the seed laser trajectory in the modulator, looking at FERMI performance as a reference.
	\subsection{The Environment of the Experiment}
	\begin{figure}[htbp]
		\centering
		% \documentclass{standalone}
% \usepackage{tikz}

% \begin{document}
    \newcommand{\boundellipse}[3]% center, xdim, ydim
{(#1) ellipse (#2 and #3)}

    \begin{tikzpicture}[scale=0.75, transform shape]
        \draw [line width=1] (-2.5,0.9) -- (-1.5,0.9) -- (-1.5,1.1) -- (-2.5,1.1) -- (-2.5,0.9);
        \node (source) at (-2,1.5) {laser source};
        \draw [red,thick](-1.5,1) -- (-1,1);
        \draw [red,thick,dashed](-1,1) -- (-0.5,1);
        \draw [red,thick](-0.5,1) -- (0,1);
        \draw [line width=1] (-0.5,1.5) -- (0.5,0.5); % TT1
        \draw (-0.2,1.2) -- (-0.1,1.3) -- (0.3,0.9) -- (0.2,0.8);
        \node (tt1) at (0.5,1.5) {TT1};
        \draw [red,thick](0,1) -- (0,0);
        \draw [red,thick,dashed](0,0) -- (0,-1);
        \draw [red,thick](0,-1) -- (0,-2);
        \draw [line width=1] (-0.5,-1.5) -- (0.5,-2.5); % TT2
        \draw (-0.2,-1.8) -- (-0.3,-1.9) -- (0.1,-2.3) -- (0.2,-2.2);
        \node (tt2) at (-0.5,-2.5) {TT2};
        \draw [red,thick](0,-2) -- (0.5,-2);
        \draw [red,thick,dashed](0.5,-2) -- (1,-2);
        \draw [red,thick](1,-2) -- (2,-2);
        \draw [line width=1.5, dashed] (1.5,-3) -- (1.5,-1) node[above]{Screen/CCD1};
         \draw [line width=1] (2,-2.5) -- (2,-1.5) -- (4,-1.5) -- (4,-2.5) -- (2,-2.5);
         \draw \boundellipse{2.75,-2}{0.25}{0.05}[fill=blue,opacity=0.75];
         \node (ebeam) at (2.75,-2.2) {\footnotesize{$e^-$ beam}};
         \node (tt1) at (3,-1.2) {modulator};
         \draw [red,thick,dashed] (2,-2) -- (4,-2);
        \draw [red,thick](4,-2) -- (4.5,-2);
        \draw [line width=1.5, dashed] (4.5,-3) -- (4.5,-1) node[above]{Screen/CCD2};
        \draw [black, thick] (4.75,-2) -- (5,-2);
        \draw [black, thick] (5.25,-2) -- (5.5,-2);
        \draw [black, thick] (5.75,-2) -- (6,-2);
        \draw [black, thick] (6.25,-2) -- (6.5,-2);
        \draw [black, thick, ->] (6.75,-2) -- (7,-2);
        \node (lg1) at (5.875,-1.25) {generated};%{light};
        \node (lg2) at (5.875,-1.75) {photon beam};%{generation};
        \draw [line width=1] (7.25,-3) -- (7.75,-3) -- (7.75,-1) -- (7.25,-1) -- (7.25,-3);
        \node (source) at (7.5,-0.75) {$I_0$ monitor};
    \end{tikzpicture}
    
% \end{document}
		\caption{A schematic view on the set-up of the FERMI FEL.}
		\label{fig:schematic_FEL}
	\end{figure}
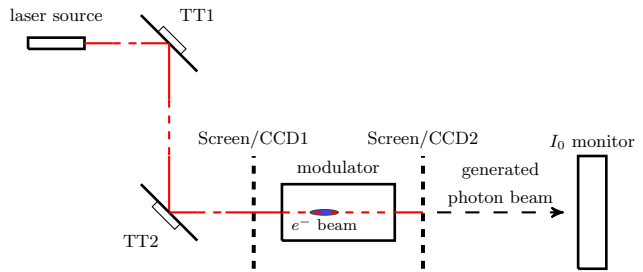
	%	The most critical parameters in a seeded free-electron laser is the temporal and transverse overlap of the electron and laser beams in the modulator magnetic section. 
	%	The problem is simplified by keeping constant the trajectory of the electron beam and the mechanical delay line that controls the temporal alignment.
	A schematic overview of the set-up is provided in Fig.~\ref{fig:schematic_FEL}.
	Two mirrors, TT1 and TT2, are used to control the trajectory of the laser by tilting and inclining, which gives a total of four degrees of freedom (DOF). In turn, the laser overlaps with the electron beam between the two removable screens, CCD1 and CCD2. Lastly, the monitor measures the intensity, $I_0$.\\
	The final problem faced consists of optimising the seed laser trajectory to match the electron beam and, consequently, increasing the FEL radiation intensity.
	The RL algorithm accesses the problem via an \emph{openai gym} environment \cite{Brockman2016}, which interfaces the RL algorithm with the control system.\\
	The state $\bs$ is a four dimensional vector holding the current voltage values applied to the piezo motors, where each component lies within the normalized interval [0, 1]. Two values correspond to the two DOF of each mirror. The action $\ba$ is four dimensional, namely a delta step on the voltages at time-step $t$:
	\begin{align}
		\bs_{t+1} = \bs_t+\ba_t,
	\end{align} 
	and is limited by $ |\ba_t|\leq \ba_\text{max}=1/12$ \footnote{The value was determined experimentally to guarantee reproducibility.}.
	The training is done in episodes with a uniformly \footnote{Sampled in the normalised state-space.} randomised initial state and a specific maximal length of taken steps afterwards - the horizon. The horizon during the training is 500 steps. An early termination happens if a specific threshold is obtained, which lies at 95\% of the maximal achievable intensity of the system.\\
	During the training phase exploration and the training of the function, approximator takes place.
	After a defined number of training episodes, verification of 50 or 100 episodes is performed, where the algorithm only applies the learned policy.
	\section{Deep Reinforcement learning}
	Generally, two main categories of deep RL algorithms can be distinguished: model-free RL (MFRL) and model-based RL (MBRL). MFRL algorithms learn directly from the interaction with a system, while MBRL algorithms learn a dynamics model from these interactions first. Despite the success of MFRL in several domains such as robotics and video games \cite{Heess2017, Schulman2017,Silver2014,Lillicrap2015,OpenAI2018}, their applications is mainly restricted to simulated environments due to their low sample-efficiency. However, putting constraints on the algorithms, as discussed in Section~\ref{s:Model-free reinforcement learning}, partly cures the sample inefficiency, turning them into practical tools for real-world problems.\\
	On the other hand, MBRL shows a much better sample efficiency. The challenge for these methods is to learn an accurate model, which turns out to be rather hard and leads to the so-called model-bias problem \cite{Deisenroth2011}. In Section~\ref{ss:Uncertainty aware DYNA-style reinforcement learning} we discuss attempts to capture the model uncertainty to alleviate this obstacle by making use of model ensembling techniques or Bayesian neural networks resulting in algorithms, exhibiting a high sample-efficiency and a low model-bias.\\
	In what follows, we denote the set of all states $\bs$ by $\states$ and the set of all actions $\ba$ by $\actions$. Relative to state and action vectors, $(\bs,\ba)$, we define a reward function $\reward : \states \times \actions \rightarrow \real$, a scalar $\discount \in (0,1]$ called discount factor and an initial state distribution $\initstate$. $\transitions(\bs_{t+1}|\bs_t,\ba_t)$ characterises the probability to end in a state $\bs_{t+1}$ if an action $\ba_t$ is taken at state $\bs_t$. The tuple \mbox{$\mdp = (\states,\actions,\transitions,\initstate,\reward,\discount)$} defines a Markov decision process.\\
	The mapping $\policy(\ba_t|\bs_t):\bs_t\mapsto \ba_t$ is called policy and draws trajectories $\traj := (\bs_0, \ba_0, \bs_1, \ba_1,\dots \bs_\horizon,\ba_\horizon)$ with a horizon $\horizon$.
	The goal of reinforcement learning is to find a policy $\policy^*$ , which is the solution of:
	\begin{equation}
		J(\policy^*)=\max J (\policy)  = 
		\mathbb E_{\traj \sim p_{\policy}(\tau)}\left[\sum_{t=0}^{\horizon}\discount^t r(\bs_t,\ba_t)\right],\label{eq:cumulative_reward}
	\end{equation}
	where
	\begin{equation}
		p_\policy(\tau) = \initstate(\bs_0)\prod_{t_0}^\horizon\policy(\ba_t,\bs_t)\transitions(\bs_{t+1}|\bs_t,\ba_t)
		\label{eq:trajectory_distribution}
	\end{equation}
	
	is the distribution of all trajectories $\tau$.

	\subsection{Model-free Reinforcement Learning}\label{s:Model-free reinforcement learning}
	In the modern field of RL one differentiates if the policy $\policy(\ba)\approx \policy_\phi(\bs)$ or the state-action value function $Q(\bs,\ba) \approx Q_\theta(\bs,\ba)$ is approximated using a high capacity function approximator, as e.g. a deep neural network.\\
	In the first case $\policy$ is optimized directly and one speaks about policy gradient methods \cite{Sutton2018, Williams1992,Baxter2011,pmlr-v28-levine13, Schulman2015,Schulman2017}. To train these methods, a large number of direct interactions with the system is required, since they belong to \emph{on-policy} algorithms. Hence, we rule them out for direct online training but make use of their characteristics in MBRL (Section~\ref{ss:critical_design}).\\
	In the second case, we speak about Q-learning, which belongs to approximate dynamic programming. One advantage is the possible usage of use previously stored data as we use it in an \emph{off-policy} fashion, which generally needs fewer data to be trained than \emph{on-policy} methods. A combination of both is also used, known as actor-critic methods \cite{Szepesvari2010,Lillicrap2015,Silver2014}.\\
	As already adduced, MFRL algorithms learn directly from interactions with the system. An exhaustive overview can be found, e.g. in \cite{Sutton2018,Levine2020}. In the following, we discuss the normalised advantage function (\emph{NAF}) algorithm \cite{Gu2016} in some detail, as it has good characteristics for accelerator control problems. It is highly sample-efficiency, and its representational power is sufficient for most common cases\cite{Kain2020,Hirlaender2020a}. Modifications to increase the stability are subject to Section~\ref{ss:Normalized advantage function}.
	\subsubsection{Approximate dynamic programming}
	The state value function $V^\policy(\bs)$ tells us how good a state in terms of the expected return is following a specific $\policy$:
	\begin{align}
		V^\policy(\bs_t) := \mathbb E_{\traj \sim p_{\policy}(\traj|\bs_t)}\left[\sum_{t=t'}^{\horizon}\discount^{(t'-t)} r(\bs_t,\ba_t)\right].\label{eq:state-value-function}
	\end{align}
	The state-action-value function - short Q function - $Q^\policy(\bs,\ba)$, which tells us how good an action $\ba$ in a state $\bs$ following $\policy$ is, in terms of the expected return:
	\begin{align}
		Q^\policy(\bs_t,\ba_t) := \mathbb E_{\traj \sim p_{\policy}(\traj|\bs_t,\ba_t)}\left[\sum_{t=t'}^{\horizon}\discount^{(t'-t)} r(\bs_t,\ba_t)\right],
	\end{align}
	is expressed as $Q^\policy_\theta(\bs,\ba)$, where $\theta$ denotes the parameters of the function approximator, such as the weights of a neural network. By satisfying the Bellmann-optimality equation $Q^\policy_\theta$ can be trained towards the optimal $Q^*(\bs,\ba)$ by minimizing the Bellman error \footnote{$\vec{Q}_\theta$ denotes the Q-function $Q_\theta$ represented as a vector of length $|\states|\times |\actions|$.}:
	\begin{align}
		\min_\theta \left(\vec Q_\theta - \bellman^*\vec Q_\theta\right)^2.\label{eq:minimize_bellmann_optimality}
	\end{align}
	$\policy$ can be calculated via:
	\begin{align}
		\policy_\theta(\ba_t|\bs_t)=\delta(\ba_t-\arg\!\max_\ba Q_\theta(\bs_t,\ba)).
	\end{align}
	The Bellman operator $\bellman^*$ has a unique fixed point but is non-linear, due to the involved $\max$ - operator \cite{Sutton2018}:
	\begin{multline}
		\bellman^*\vec Q_\theta(\bs_t,\ba_t) := \\r(\bs_t,\ba_t)+\gamma\max_\ba\left( Q_\theta(\bs_{t+1},\ba) - Q_\theta(\bs_t,\ba_t)\right).
	\end{multline}
	The nature of this equation can cause overestimation and other complications, when using a function approximator and several attempts exist to overcome its difficulties \cite{Hasselt2015,Mnih2013,Lillicrap2015,Gu2016,Wang2015}.
	\subsubsection{Design decisions for MFRL}\label{ss:Normalized advantage function}
	One way to avoid the sensitivity to the mentioned complications is to choose a simple analytical form with an explicitly calculable maximum of the $Q$-function.
	If a specific quadratic form of the Q-function is assumed \cite{Gu2016}:
	\begin{eqnarray}
		Q_\theta(\bs,\ba) &&= \nonumber\\  & & -\frac{1}{2}(\ba-\mu_\theta(\bs))P_\theta(\bs)(\ba-\mu_\theta(\bs))^T+V_\theta(\bs).\label{eq:state-action-value-approxiation}
	\end{eqnarray}
	$P_\theta(\bs)$ is a state-dependent, positive-definite
	square matrix, parametrised in a specific way \cite{Gu2016}.
	One modification, which we made in our experiments is the application of a twin network (weights for network $i$ denoted by $\theta^i$). Only one network is used to obtain the policy, while the other is employed for the update rule to avoid over-estimation. It is motivated by double Q-learning \cite{NIPS2010_091d584f,Hasselt2015,fujimoto2018addressing}.
	The maximum is given analytically as $\max_\ba Q(\bs,\ba) = V(\bs)$, hence from Eq.~\ref{eq:minimize_bellmann_optimality} the loss $\mathcal L$ can be formulated as:
	\begin{eqnarray}
		\mathcal L(\theta)=\Big( (\reward(\bs_t,\ba_t)&+&\gamma \min_{1,2} V_{\theta^i_\text{targ}}(\bs_{t+1})
		\nonumber \\
		& &-(1+\gamma) Q_\theta(\bs_t,\ba_t)\Big)^2.
	\end{eqnarray}
	$\theta_\text{targ}$ are the weights of a target network, which is softly updated \cite{Lillicrap2015, Gu2016,Silver2014}. To further stabilize the network training a small clipped artificial noise is added to the actions called \emph{smoothing} \cite{fujimoto2018addressing}. The effect is demonstrated in Appendix~\ref{appendix:naf2} on a classical example from control theory, the \emph{inverted pendulum}. In Fig.~\ref{fig:comparsion_noise} its role to increase the noise robustness can be seen. Therefore \emph{smoothing} was applied in all MFRL experiments.\\
	Our implementation has high sample-efficiency and noise resistivity. We use it as the baseline for the FERMI-FEL control problem, as it yields good results. Additional changes to the original proposal \cite{Gu2016} and a previous implementation used for accelerator control using to a prioritized replay buffer \cite{Hirlaender2020a} are discussed in Appendix~\ref{appendix:naf2}. We refer to our modified implementation as \emph{NAF2} and the implementation close to the original, as in \cite{Gu2016}, as \emph{NAF}.
	%\subsection{New NAF network architecture}
	
	%\begin{algorithm}[ht]
	%\caption{On-policy policy gradient with Monte Carlo estimator \label{alg:pg}}
	%\begin{algorithmic}[1]
	%\State initialise $\theta_0$
	%\For{iteration $k \in [0, \dots, K]$}
	%\State sample trajectories $\{\traj_i\}$ by running $\policy_{\theta_k}(\ba_t|\bs_t)$ \Comment{each $\traj_i$ consists of $\bs_{i,0},\ba_{i,0},\dots,\bs_{i,H},\ba_{i,H}$}
	%%\State compute $\return_{i,t} = \sum_{t'=t}^H \discount^{t'-t} \reward(\bs_{i,t},\ba_{i,t})$
	%%\State fit $b(\bs_t)$ to $\{\return{i,t}\}$ \Comment{use constant $b_t = \frac{1}{N}\sum_i \return{i,t}$, or fit $b(\bs_t)$ to $\{\return{i,t}\}$}
	%%\State compute $\hat{A}(\bs_{i,t},\ba_{i,t}) = \return_{i,t} - b(\bs_t)$
	%%\State estimate $\nabla_{\theta_k} J(\policy_{\theta_k}) \approx \sum_{i,t} \nabla_{\theta_k} \log \policy_{\theta_k}(\ba_{i,t} | \bs_{i,t}) \hat{A}(\bs_{i,t},\ba_{i,t})$
	%%\State update parameters: $\theta_{k+1} \leftarrow \theta_k + \alpha \nabla_{\theta_k} J(\policy_{\theta_k})$
	%%\EndFor
	%\end{algorithmic}
	%\end{algorithm}
	
	\subsection{Uncertainty Aware DYNA-style Reinforcement Learning}\label{ss:Uncertainty aware DYNA-style reinforcement learning}
	%The original DYNA algorithm \cite{Kurutach2018} is modified here in several aspects.
	
\begin{figure}[htbp]
		\centering
		\includegraphics*[width=0.4\textwidth]{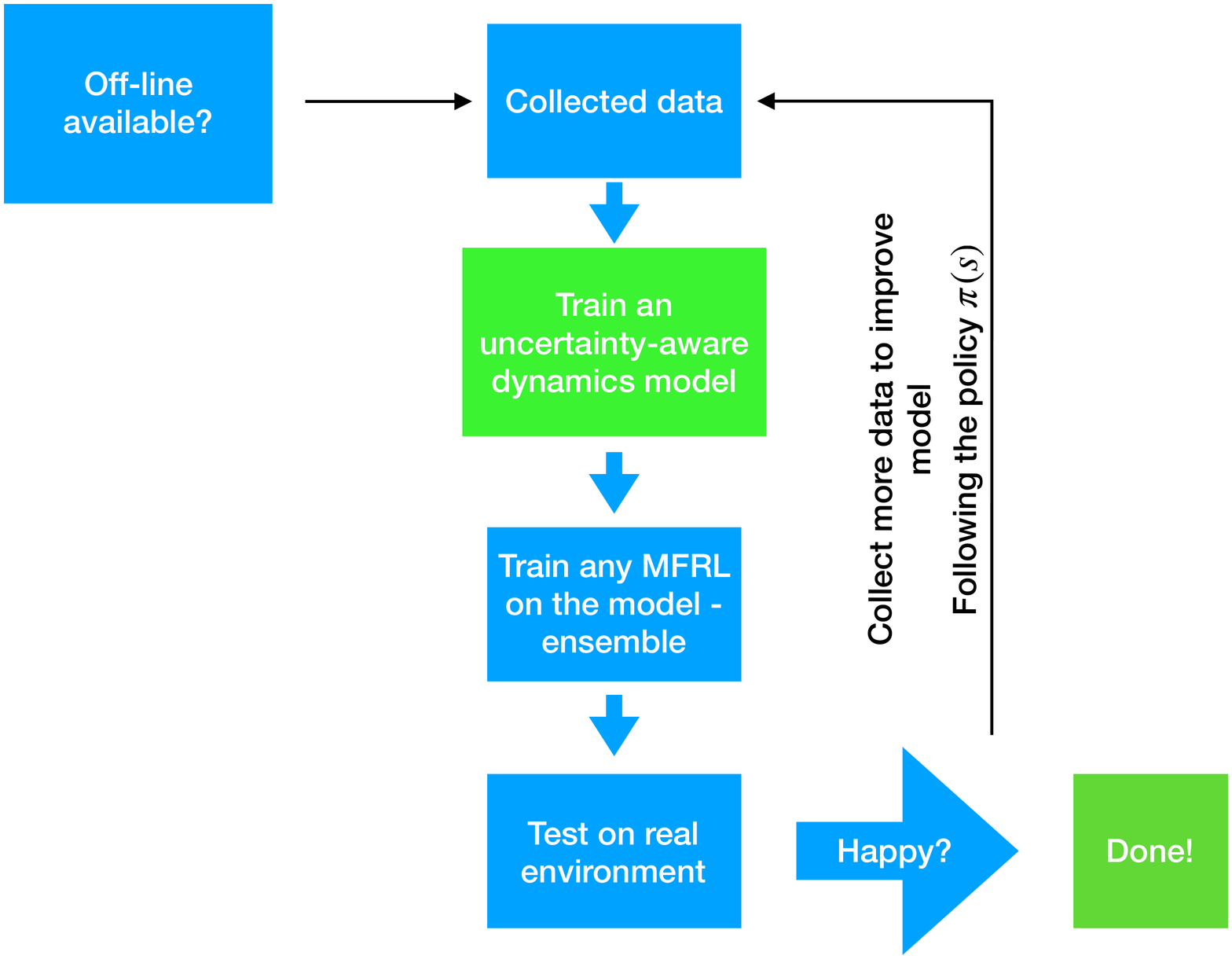}
		\caption{A schematic overview of the \emph{AE-DYNA} approach used in this paper.}
		\label{fig:MBRL_overview}
	\end{figure}

	An excellent overview of the state of the art MBRL methods is provided in \cite{Wang2019}. Suitable methods are \cite{Gal2016,6654139}, based on Gaussian processes and Bayesian Neural Networks and back-propagation in the dynamics model or \cite{Chua2018,Wang2019a}, using highly efficient sample-based methods (e.g. the cross entropy method \cite{Boer2005}). We focus on model-based data generation.\\
	In MFRL the dynamics $\transitions$ of an environment is learned implicitly, while in model-based reinforcement learning MBRL $\transitions$ is learned explicitly and approximated as $\hat \transitions$. In addition to the next state, the reward is included in the model \footnote{In some applications $\reward$ can be deduced from the current state or the state change. In such a situation, the modelling can be simplified to model directly $\hat\transitions$.}:
	\begin{align}
		\hat{f_\theta}(\bs_t,\ba_t) := \{\hat\transitions_\theta(\bs_t,\ba_t),r_\theta(\bs_t,\ba_t)\}.\label{eq:dynamics_model}
	\end{align}
	Generally, DYNA style algorithms \cite{Sutton1991} denote algorithms, where an MFRL algorithm is trained on purely synthetic data from an approximate dynamics model $\hat \transitions$ or a mixture of synthetic and real data. We use only synthetic data to reduce the interaction with the real environment to a minimum.\\
	A schematic overview of our used method is shown in Fig.~\ref{fig:MBRL_overview}. Initially, data is gathered by interaction with the real environment or read from a previous collection. An uncertainty aware model of the dynamics is trained, using `anchored ensembling' \cite{Pearce2018} on the data (details provided in Appendix~\ref{appendix:The impact of noise}), which allows taking the aleatoric (measurement errors) as well as the epistemic uncertainty (lack of data) into account. It was motivated by the fact that pure epistemic ensemble techniques as \cite{Kurutach2018} were reported to be very sensitive to aleatoric noise \cite{Wang2019}. Alternatives as \cite{Chua2018,Janner2019,Wang2019a} also take these uncertainties into consideration using probabilistic ensembles. Our implementation is simpler while showing excellent performance in considered experiments.\\
	Subsequently, an MFRL algorithm is deployed to learn the controller on the dynamics model by only using the synthetic data and in-cooperating the model uncertainty, as explained in Section~\ref{ss:The uncertainty aware dynamics model}.\\
	After a defined number of training steps, called an epoch, the controller is validated on each model of the ensemble, individually. If there is no improvement in a number smaller than the total number of the models, the controller might be tested on the real environment for several episodes. In this way, `over-fitting' on a wrong model is avoided, as discussed in \cite{Kurutach2018}. The training is stopped if the controller produces satisfying results on the real environment. Otherwise, a batch of new data is collected, and the model is improved, and consecutively a new epoch starts.
	\subsection{Critical Design Decisions in MBRL}\label{ss:critical_design}
	In the following, the most important aspects of a successful application of MBRL are discussed. To our knowledge uncertainty aware MBRL is applied for the first time with success on an accelerator problem and might be, despite its incredible potential, demanding. We made some algorithmic design decisions, which we think are beneficial for problems typically found in accelerator operation.\\
	%	 For the DYNA-AE algorithm, the important points are:
	%	\begin{itemize}
		%		\item The ANN - including the prior - number of models - noise level - early stopping.
		%		\item The number of data-points and the policy.
		%		\item The uncertainty - how is this included?
		%		\item The episodic design - avoid long trajectories.
		%		\item The MFRL agent as, e.g. the TRPO and PPO or the SAC and its training.
		%		\item Tuning the algorithm on a model.
		%	\end{itemize}
	%\section{Proof-of-principle application of RL Agent at FERMI trajectory correction}
	%\NB{Niky}
	\subsubsection{The uncertainty aware dynamics model}\label{ss:The uncertainty aware dynamics model}
	The 'approximately Bayesian Ensembling' method usually yields good results already with a small number of models. Empirical results showed that already three models were satisfactory to see definite improvements over a single network approach (see Fig.~\ref{fig:Compare_models_sizes}). The main goal is not to determine the exact posterior probability of the learned model but, as already mentioned, to provide the uncertainty to not over-train the controller at areas without sufficient data.\\
	A small densely connected two-layer network with around 15-25 nodes each and $tanh$ activation functions was used. The last layer was linear, and the inputs $\{\bs,\ba\}$ were normalised to the interval [-1,1].\\
	The prior is controlled by the initialisation of the weights of the network. It defines how large the variation of the approximated function is expected, and they were randomised uniformly in an interval $[-\Delta, \Delta]$, where $|\Delta|\leq0.1$. The weights of the last layer were normalised by the number of its nodes. For the moment, only homoscedastic Gaussian errors are considered with a standard deviation $\sigma_\epsilon$. This is respected in a regularisation term (details see Appendix~\ref{appendix:The impact of noise}) and significantly improves the training in the presence of noise, as shown in Fig.~\ref{fig:comparsion_noise_ae_dyna}. \\
	Several methods of network training were implemented. \emph{Early stopping} \cite{Goodfellow2016}, with a learning rate of $10^{-3}$ to $10^{-4}$  was employed with a waiting time of about 20 steps and a validation ratio of 20\%. The number of training steps is increasing with more data in close-by areas. It is followed by a shrinking uncertainty at regions where no data is available, leading to better training performance. A fixed maximal loss threshold was also beneficially tested. In the experiments, a combination of both was taken.
	\subsubsection{The controller algorithm}
	To decide which MFRL algorithm to be trained on the learned model, two already mentioned main algorithm classes are considered  (Section~\ref{s:Model-free reinforcement learning}): \emph{on-} and  \emph{off-policy} algorithms \cite{Sutton2018}. \emph{On-policy} algorithm show a stabler and monotonic convergence to a local maximum in general, while \emph{off-policy} algorithms have the advantage that they might converge to the global maximum.\\
	The \emph{on-policy} algorithm trust region policy optimisation \emph{TRPO} \cite{Schulman2015} provides the theoretical guarantee of a monotonic improvement of the policy. We use this algorithm in experiments labelled as \emph{ME-TRPO}, which stands for model-ensemble TRPO as proposed in \cite{Kurutach2018}. As reported in \cite{Kurutach2018}, using proximal policy optimisation \cite{Schulman2017}, another prominent \emph{on-policy} algorithm, reveals minor performance, which was confirmed in our tests. \\
	An attractive \emph{off-policy} algorithm is the soft-actor-critic (\emph{SAC}) \cite{fujimoto2018addressing,Hill2018}. \emph{SAC} not only tries to maximise Eq.~\ref{eq:cumulative_reward} but also simultaneously is regularised to maximise the entropy in the action space \cite{Haarnoja2018a}. In this way, exploration is encouraged to avoid getting stuck in a local optimum, and a good trade-off between exploration and exploitation is achieved. We refer to experiments using the \emph{SAC} as \emph{AE-DYNA-SAC}. 
	For \emph{AE-DYNA-SAC} the controller was reset after each re-training of the dynamics model to profit from its exploration features, while for \emph{ME-TRPO} the policy was improved continuously to exploit its monotony.
	\subsubsection{Handling of the model uncertainty}
	One of the most crucial points is how the MFRL controller treats the model uncertainties. Several different approaches were studied. The approximate posterior probability is obtained by taking the mean $\mu_m$ and the standard deviation $\sigma_m$ of the models and sample from $\gauss(\mu_m,\sigma_m)$. Another strategy is to randomly select a specific model out of the ensemble at each training step, which demonstrated the best performance in our \emph{ME-TRPO} experiments. This was also reported in the original implementation of the \emph{ME-TRPO} algorithm \cite{Kurutach2018}. In a pessimistic setting, one would only select the model resulting in the lowest predicted reward (as used, e.g. in \cite{kidambi2020morel}) and showed monotonic but too slow improvement for our online training.
	Good results were also obtained by following a randomly selected single model for each full episode, which was used in the \emph{AE-DYNA-SAC}. 
	\subsubsection{The data acquisition}
	The number of data-points in a new batch to improve the model has to be chosen carefully. It is influenced by the number of randomly collected initial data-points (and then by the used policy $\policy$). The initialisation phase has to be selected not too small to minimise the risks of getting trapped in a local minimum for too long. Afterwards, from our experience, at least one full episode per epoch should be taken. We decided to use a short horizon to diminish the impact of the compound error \cite{Janner2019}, the accumulation error following a wrong model, as well known in \emph{DYNA}-style approaches. The maximal number of steps in our experimental runs per episode was ten. During the data acquisition on the real system, the latest learned policy $\policy$ was taken with some small additional Gaussian noise to improve exploration.

	\section{Experimental results from FERMI RL online tests}\label{sec:Experimental results from FERMI RL online tests}
	Several tests were performed on the FERMI FEL, as discussed in the previous sections.
	The primary purpose was to try the newly implemented algorithms on a real system to evaluate their operational feasibility. 
	Because of the tight schedule at FERMI, only a few shifts of several hours could be reserved to carry out these experiments.\\ 
	%	Since there was a positive outcome, we deduce that the utilised methods are at an operational level.\\
	Table~\ref{tab:overview_algorithms} shows a comparison of the discussed algorithms. Details are provided in the appendix. All algorithms are suitable for control tasks, as common in accelerator operation. Their strength lies in the high sample-efficiency, but they differ in representational power and noise robustness. In Section~\ref{sec:Experimental results from FERMI RL online tests} the \emph{NAF2},
	\emph{ME-TRPO} and \emph{AE-DYNA} are tested and compared on the real environment. The \emph{NAF2} algorithm, the representative for highly sample efficient MFRL algorithms, is discussed first.
	\begin{table}[htbp!]%The best place to locate the table environment is directly after its first reference in the text
		\caption{\label{tab:overview_algorithms}%
			Overview of the algorithms.
		}
		%\begin{tabular}{l| p{1.5cm}p{1.5cm}p{1.5cm}p{1.5cm}}
		\begin{tabular}{l| p{1.3cm}p{1.45cm}p{1.45cm}p{1.45cm}}
			\textrm{Algorithm}&
			\textrm{Class}&
			\textrm{Noise \newline robustness}&
			\textrm{Represent. power}&
			\textrm{Sample \newline efficiency}\\
			\hline
			\emph{NAF} & MFRL & low &low &high\\
			\emph{NAF2} &MFRL & high &low& high\\
			\emph{ME-TRPO} & MBRL & low& high& high\\
			\emph{AE-DYNA} & MBRL & high& high& high\\
		\end{tabular}
		
	\end{table}
	
	\subsection{MFRL Tests}
	\begin{figure}[htbp]
		\centering
		\includegraphics*[width=0.4\textwidth]{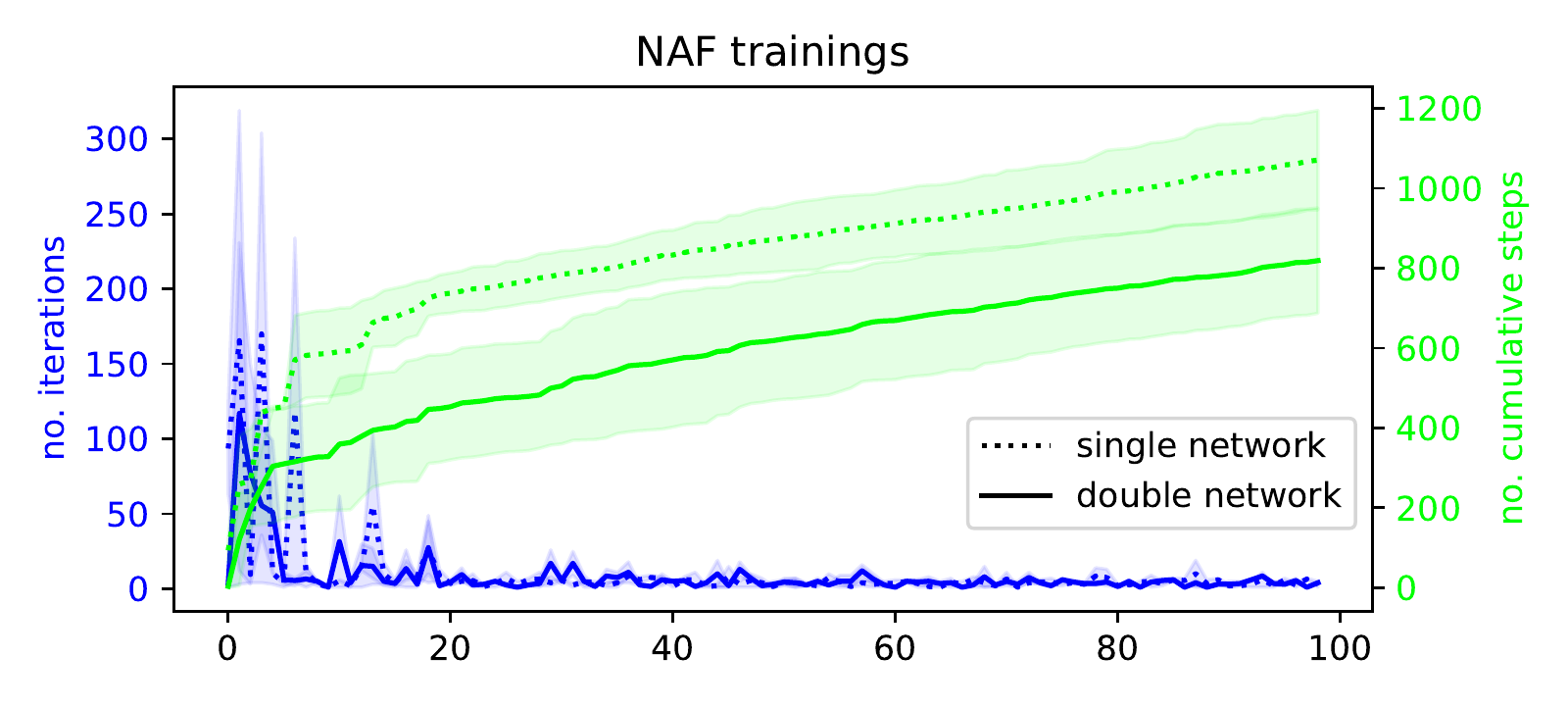}
		\caption{The training of different variants of the \emph{NAF2} algorithm  on the FERMI FEL, averaged over two complete trainings (the standard-deviations are indicated by the shaded areas). The number of iterations (blue) shows the steps until the intensity is optimised, starting from a random initial position.}
		\label{fig:NAF_training}
	\end{figure}
	\begin{figure}[htbp]
		\centering
		\includegraphics*[width=0.4\textwidth]{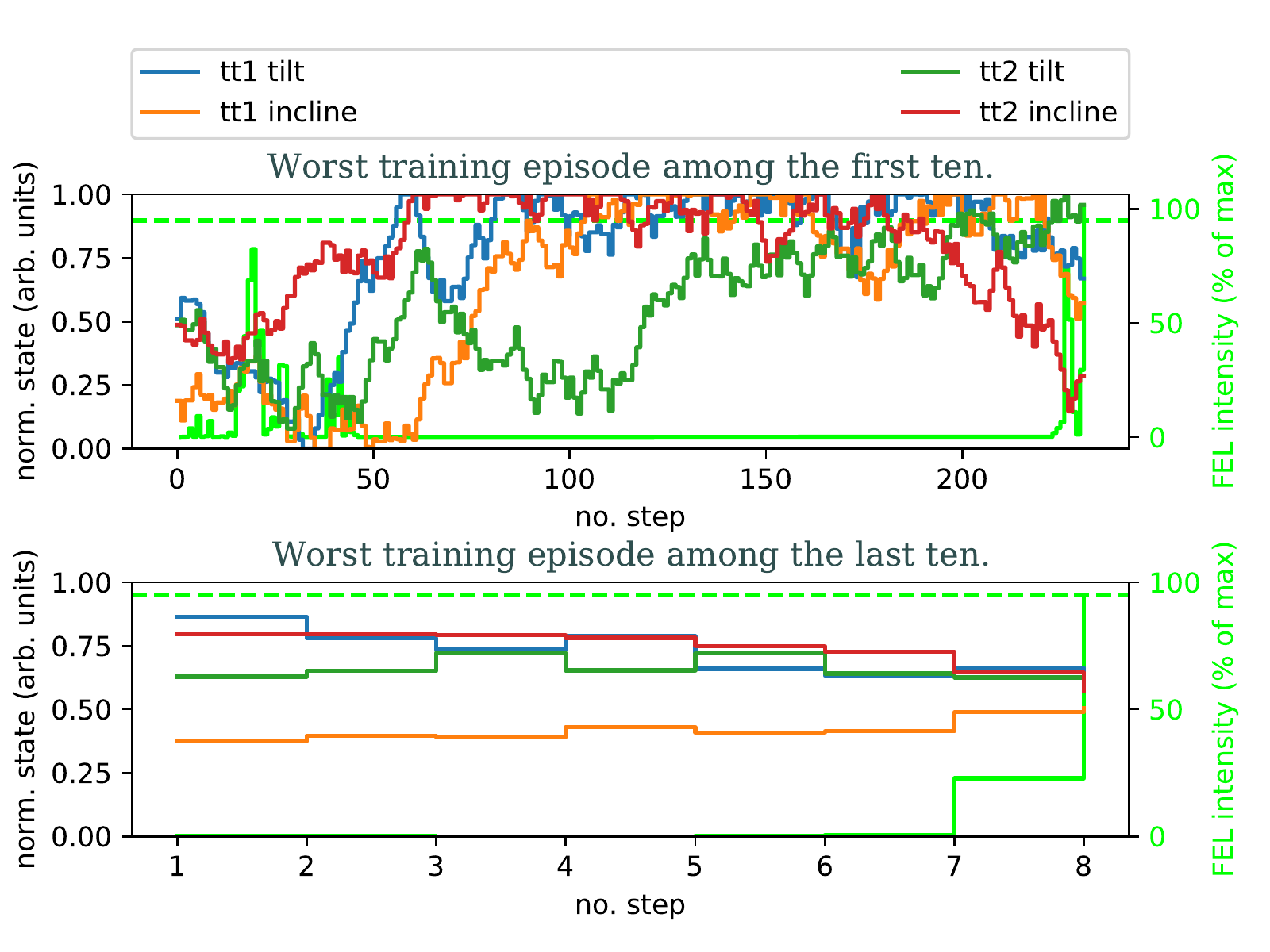}
		\caption{The evolution of the states of the \emph{NAF2} algorithm on the FERMI FEL using a double network during the training.}
		\label{fig:NAF_evolution_double}
	\end{figure}
	\begin{figure}[htbp]
		\centering
		\includegraphics*[width=0.4\textwidth]{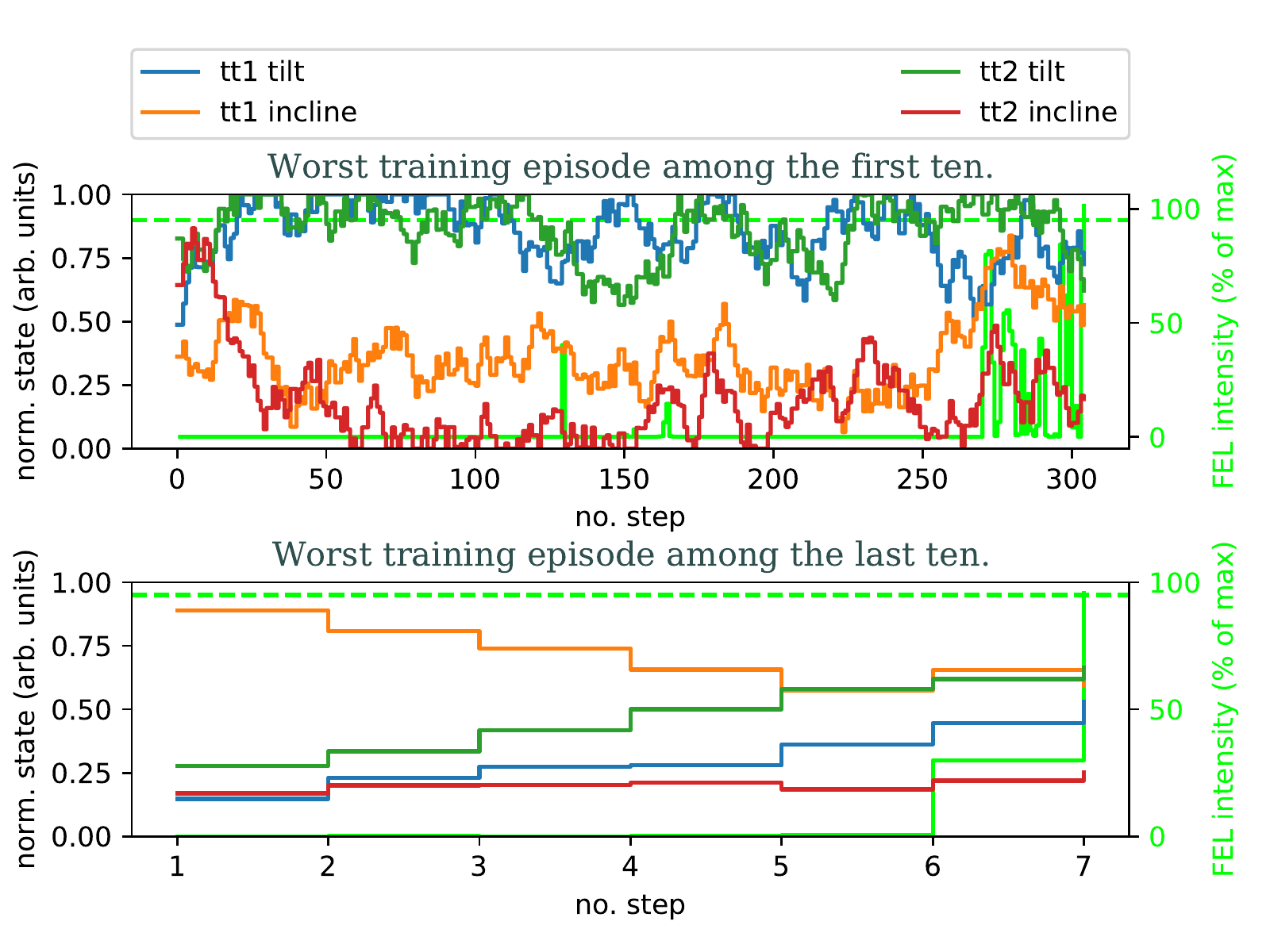}
		\caption{The evolution of the states of the \emph{NAF2} algorithm on the FERMI FEL using a single network during the training.}
		\label{fig:NAF_evolution_single}
	\end{figure}
	\begin{figure}[htbp]
		\centering
		\includegraphics*[width=0.47\textwidth]{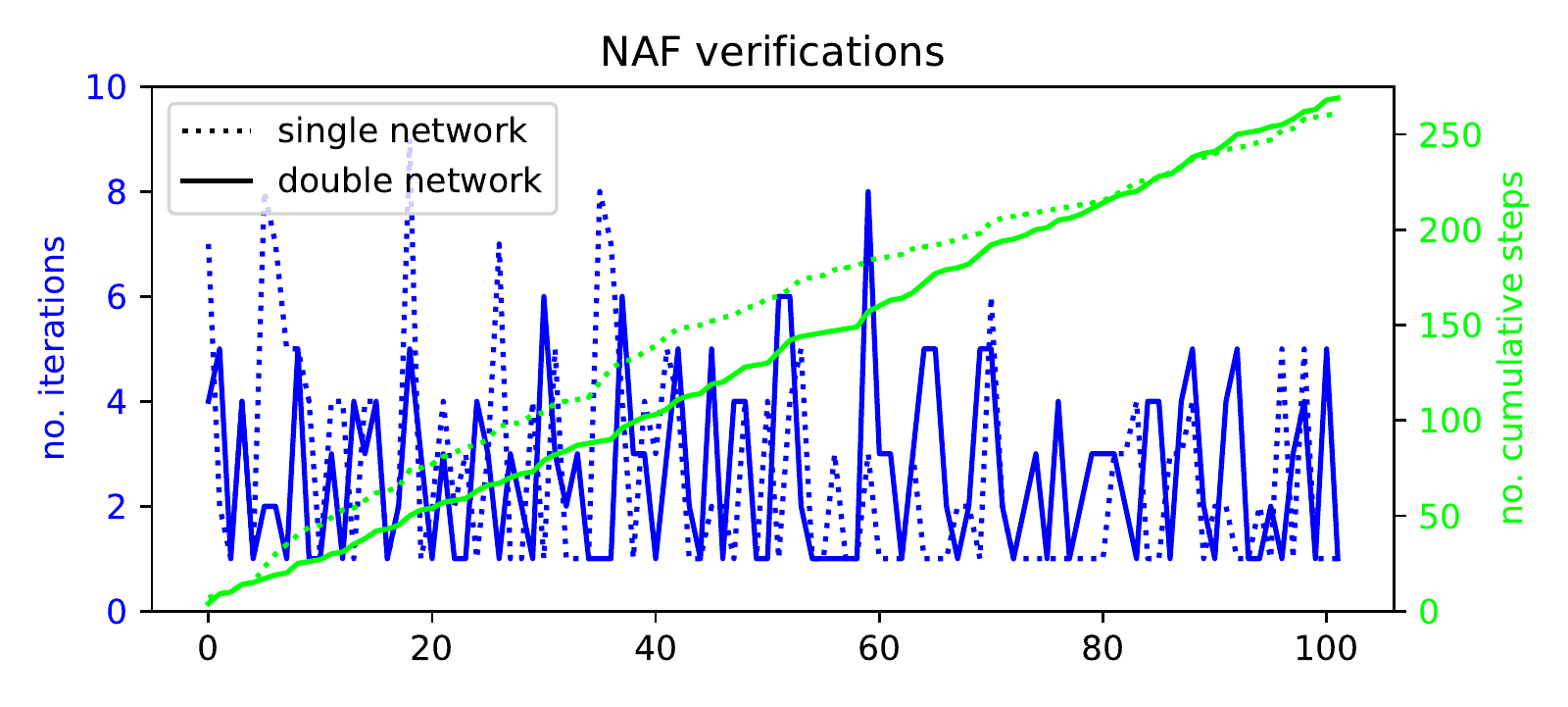}
		\caption{The verification episodes of the variants of the trained model-free \emph{NAF2} algorithm on the FERMI FEL. The number of iterations (blue) shows the steps until the intensity is optimised, starting from a random initial position.}
		\label{fig:NAF_verification}
	\end{figure}
	\begin{figure}[htbp]
		\centering
		\includegraphics*[width=0.4\textwidth]{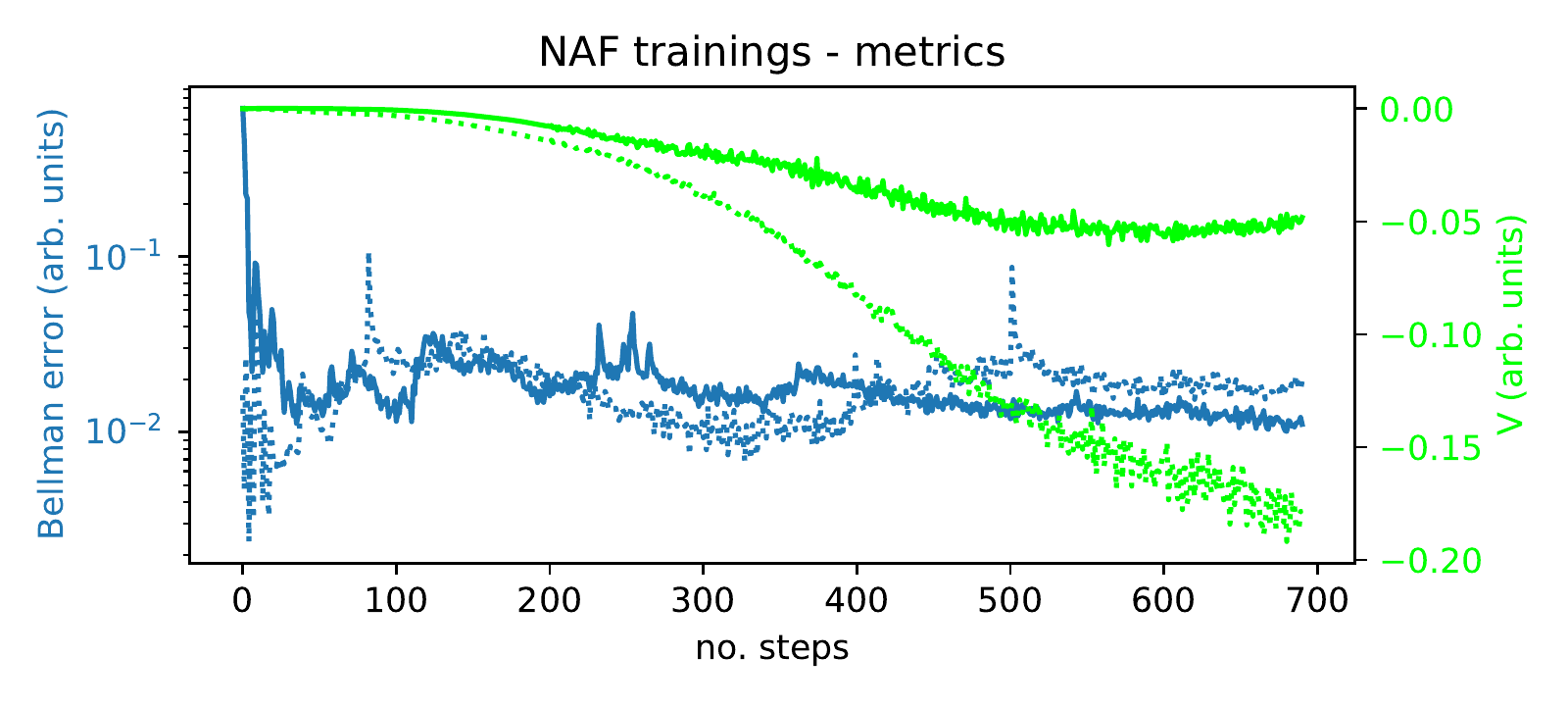}
		\caption{The training metrics of the \emph{AE-DYNA-SAC} on the FERMI FEL using a single network (dashed) and a double network (solid). The Bellman error (Eq.~\ref{eq:minimize_bellmann_optimality}) and the state-value function (Eq.~\ref{eq:state-value-function}) are shown. }
		\label{fig:NAF_convergence}
	\end{figure}
	Four tests were carried out, two using a single network and two using the double network architecture. 
	Fig.~\ref{fig:NAF_training} displays the results, averaged over the two tests \footnote{Considering the training is stochastic with random initial conditions, by chance, the training of the double network variant started above the threshold in both cases.}. A training of 100 episodes was accomplished. The number of iterations per episode is plotted, including the cumulative number of steps. During the training, in all episodes, the 95\% intensity reward threshold was surpassed; hence they were successfully finished. The evolution of the states is provided in Fig.~\ref{fig:NAF_evolution_single} and Fig.~\ref{fig:NAF_evolution_double} for the single and double network. At the boundary of the domain, it takes time to learn since many actions are mapped onto the same state \footnote{The states are clipped to $[0,1]$.}. The behaviour indicates that the double network converges faster and spends less time at the boundary.\\
	In the verification of 100 episodes, Fig.~\ref{fig:NAF_verification}, both algorithms show similar performance, while the double network needed fewer training steps and reveals a more stable overall performance. Additionally, the convergence metrics of the two algorithms is plotted in Fig.~\ref{fig:NAF_convergence} against the number of training steps. The blue curves show the Bellmann error Eq.~\ref{eq:minimize_bellmann_optimality}, which is comparable in both cases. The state-value function $V$ (Eq.~\ref{eq:state-value-function}), which is a direct output of the neural net (Eq.~\ref{eq:state-action-value-approxiation}), converges to a reasonable value for the double network within the shown 700 steps, whereas the single network seems to overestimate the value. In the single-network case, convergence is reached after around 1400 steps. In the plot, $V$ was averaged over 200 randomly selected states.
	\subsection{MBRL Tests}
	\begin{figure}[htbp]
		\centering
		\includegraphics*[width=0.4\textwidth]{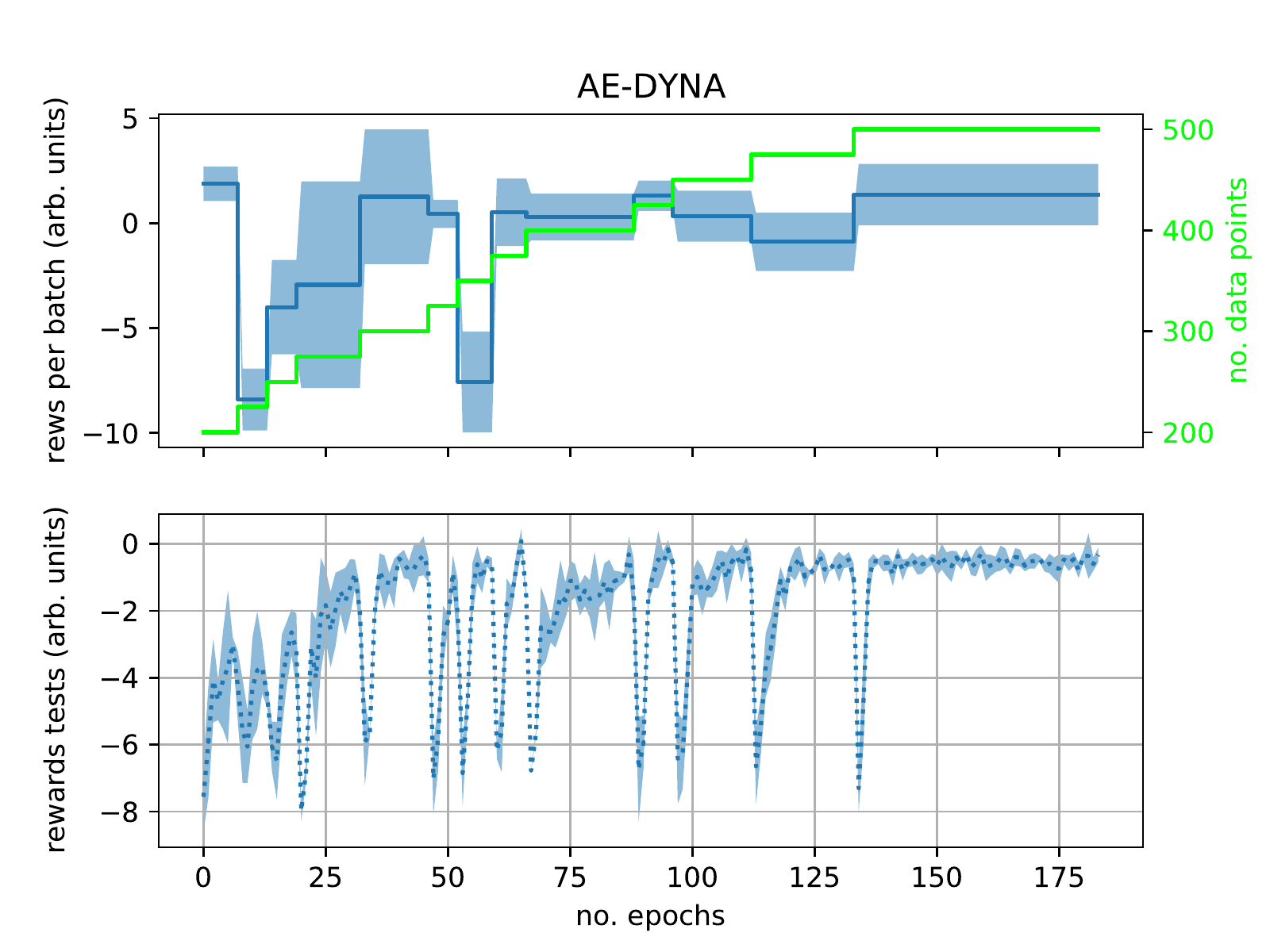}
		\caption{The training observables of the \emph{AE-DYNA-SAC} on the FERMI FEL. Detail are provided in the text.}
		\label{fig:AE-DYNA_observables}
	\end{figure}
	
	\begin{figure}[htbp]
		\centering
		\includegraphics*[width=0.4\textwidth]{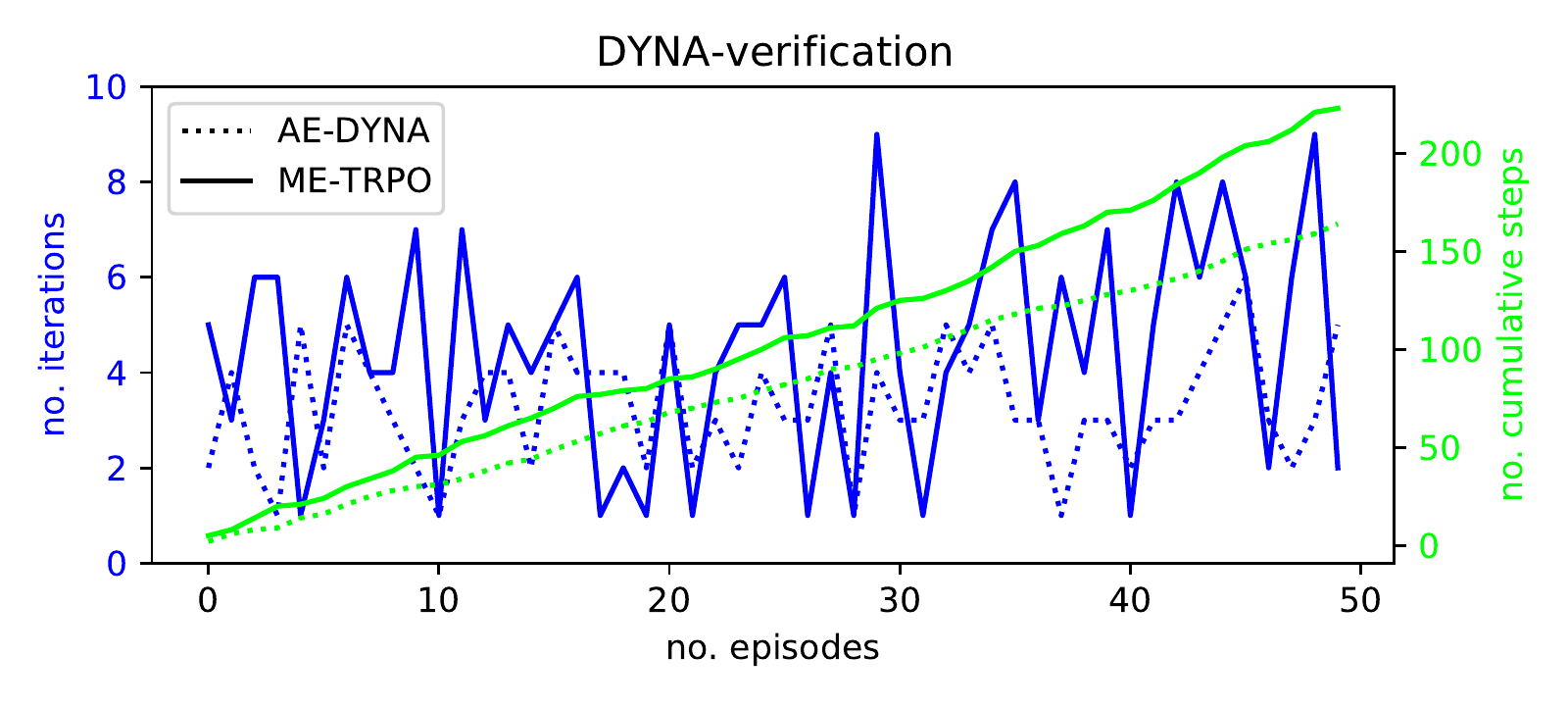}
		\caption{The verification episodes of the trained model-based methods: \emph{ME-TRPO} and \emph{AE-DYNA-SAC} on the FERMI FEL. The number of iterations (blue) shows the steps until the intensity is optimised, starting from a random initial position.}
		\label{fig:AE-DYNA_verification}
	\end{figure}

	\begin{figure}[htbp]
		\centering
		\includegraphics*[width=0.4\textwidth]{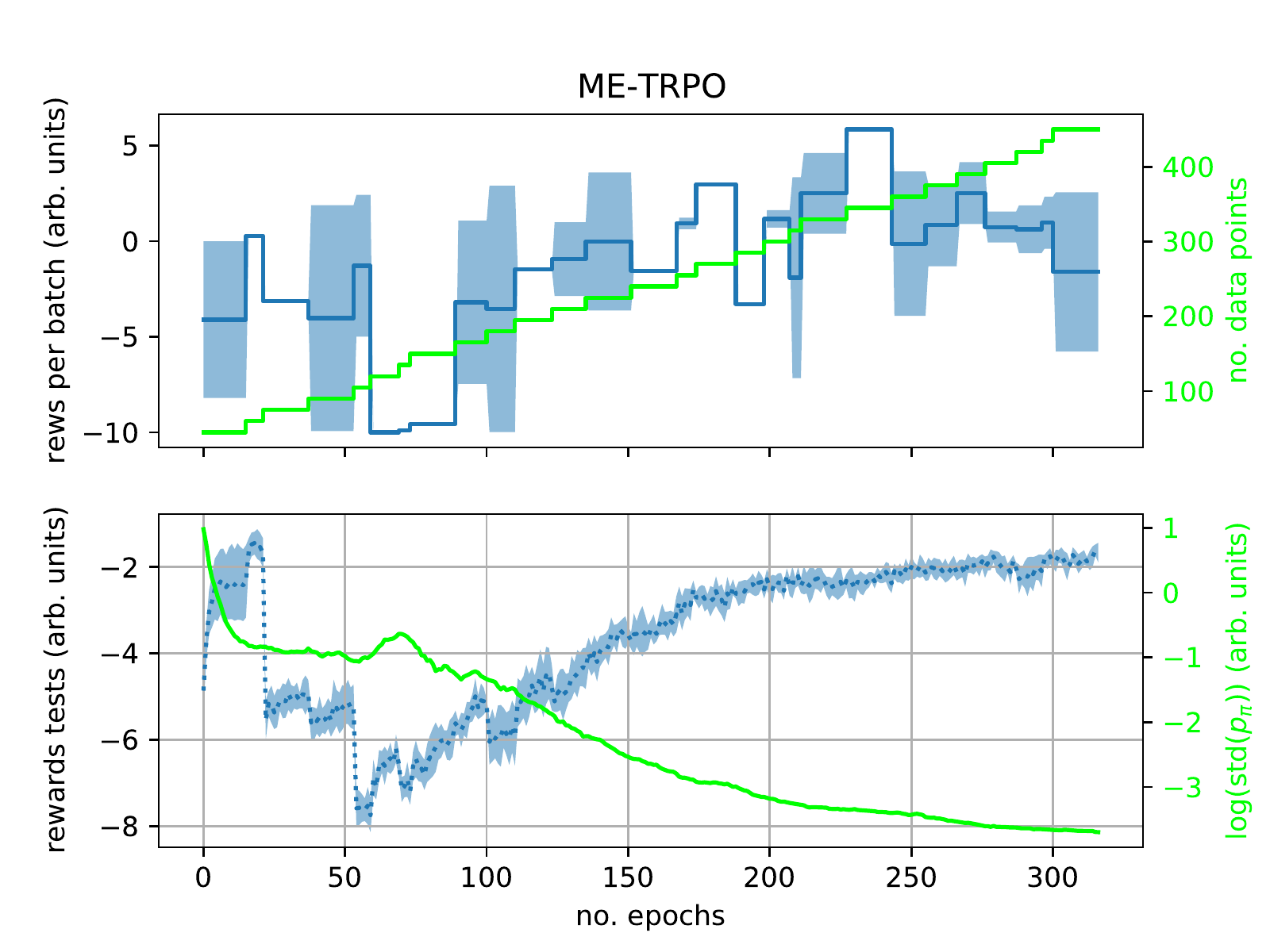}
		\caption{The training observables of the \emph{ME-TRPO} on the FERMI FEL. Details are provided in the text.}
		\label{fig:ME-TRPO_observables}
	\end{figure}
	
	\begin{figure}[htbp]
		\centering
		\includegraphics*[width=0.4\textwidth]{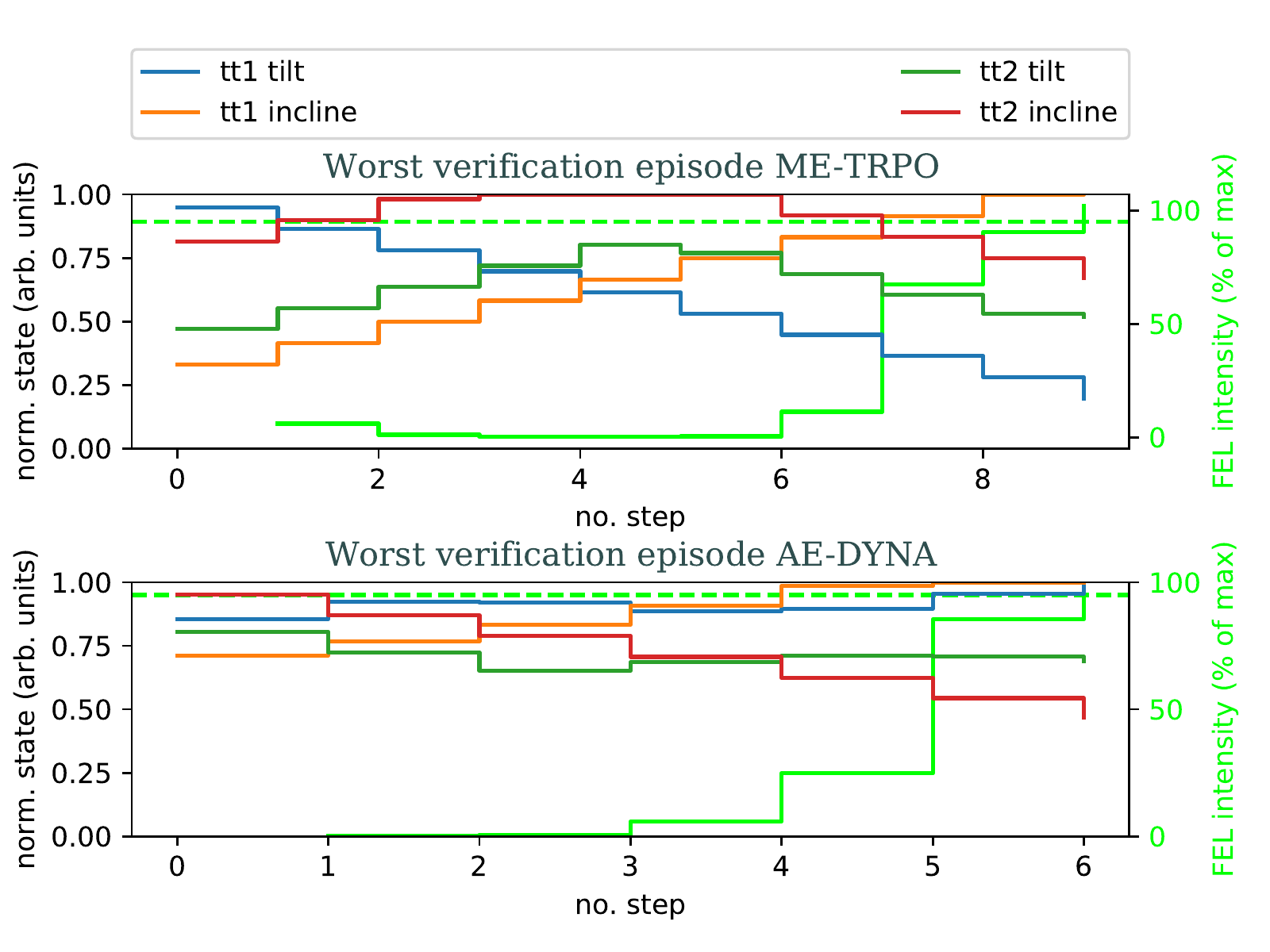}
		\caption{The evolution of the states during the worst verification episodes of the trained \emph{ME-TRPO} and the \emph{AE-DYNA-SAC} on the FERMI FEL.}
		\label{fig:Worst_episode_MBRL}
	\end{figure}
	%	\begin{figure}
		%		\centering
		%		\includegraphics*[width=0.5\textwidth]{Figures/ME-TRPO_verification.pdf}
		%		\caption{The verification of the ME-TRPO on the FERMI FEL.}
		%		\label{fig:ME-TRPO_verification}
		%	\end{figure}
	
	The second test campaign employed the \emph{AE-DYNA} algorithm as a representative for pure MBRL algorithms. As discussed, two variants were implemented: the \emph{ME-TRPO} variant and the \emph{AE-DYNA-SAC} variant. The algorithmic design details were discussed in Section~\ref{ss:critical_design}. 
	\\ To exploit the convergence properties of the \emph{TRPO}, the policy was never reset as can be seen in Fig.~\ref{fig:ME-TRPO_observables}. The upper figure shows the total reward per batch gathered from the real environment and the number of data-points used to train the dynamics model as a function of the number of epochs. In the lower plot, the average cumulative reward of ten episodes as achieved by the \emph{TRPO} on the individual models of the ensemble independence of the epochs is drawn. During an epoch, the \emph{TRPO} is trained for 10000 steps on the synthetic data. The shaded area shows the corresponding standard deviation to indicate the uncertainty of the dynamics model. The algorithm continuously improves after initial performance drops caused by the insufficiently trained model. As a measure of convergence of the \emph{TRPO}, the logarithm of the standard deviation of $p_\policy$ (Eq.~\ref{eq:trajectory_distribution}) is visualised. The training was stopped after 450 steps collecting 25 steps each dynamics training.
	\\ As shown in Fig.~\ref{fig:AE-DYNA_verification}, all of the 50 verification episodes were successfully finished after a few steps. To verify the impact of the ensemble technique, a test with a single network using the same hyper-parameters has been applied, where no convergence within 500 data-points has been observed. \\
	Secondly, the \emph{AE-DYNA-SAC} was tested. As discussed in Section~\ref{ss:critical_design}, in this test, the controller was reset each time when the model was re-trained, and consequently, the performance drops each time as shown in Fig.~\ref{fig:AE-DYNA_observables}.\\
	In contrast to the \emph{ME-TRPO} training, the data batches consisted of 50 data-points with an initial random walk of 200 steps. The number of initial steps was chosen high enough because otherwise, the convergence is slowed down enormously so that the training becomes unfeasible on a real machine (discussed in Section~\ref{ss:critical_design}). Each epoch consists of 2500 steps of controller training on the model. The training was stopped after the acquisition of 500 data-points. The verification was executed as in the first test. Again, the success of all 50 episodes is 100\%. The number of needed iterations per episode is less than for the ME-TRPO, which can be seen in Fig.~\ref{fig:AE-DYNA_verification}. The somewhat better performance might be a result of the higher number of data-points (50), but in general, this method exhibited better asymptotic performance than the \emph{ME-TRPO} variant.\\
	In both experiments, the training used a small number of data-points while still successfully solving the problem. In Fig.~\ref{fig:Worst_episode_MBRL} the worst verification episodes of both experiments are plotted. The experiments, taken on different days, may influence the boundary conditions and hence the performance. However, more extended training would increase the performance in both cases, especially for the \emph{ME-TRPO}. Fig.~\ref{fig:Worst_episode_MBRL} shows that the tt2 incline and tilt parameters for the \emph{ME-TRPO} move up and then down towards the optimal position, which a better dynamics model would improve.

	\section{Discussion and outlook}
	This paper presents the applicability of deep reinforcement learning on the optimisation of the FERMI FEL intensity. Two different approaches were tested: model-free and model-based.\\
	Both experiments yielded satisfactory results showing that a non-linear and noisy problem could be solved in a feasible number of training steps. The results of the verification are summarised in Table~\ref{tab:overview_verification}. In the model-free case, around 800-1000 and in the model-based 450-500 data-points were used. In all experiments, all verification episodes were finished successfully. \emph{NAF2} performs better in terms of episode length while the average reward in the \emph{DYNA}-style algorithms is higher. The experiments were done on different days, hence under slightly different conditions.\\
	Usage of the proposed methods in an operational way is attractive and could replace in the future the current optimisation method, which needs destructive screen measurement. Hence online retraining could be done, and valuable time could be saved.\\ 
	The MFRL methods were slightly better in the final performance, but more samples were collected during the training. The MBRL methods had to be stopped due to a lack of available experimental time. Hence additional studies regarding a long time performance would be interesting.\\ One big issue in applying MBRL methods can be the computational time to train the controller on the model. In our tests, the time to acquire data from the system was only a fraction of the time needed to train the agent. The used methods could be parallelised to reduce the computational time.
	\begin{table}[htbp]%The best place to locate the table environment is directly after its first reference in text
		\caption{\label{tab:overview_verification}%
			An overview of the verification performance of the different trained algorithms on the FERMI FEL, including their standard deviation.
		}
		\begin{tabular}{l| p{1.1cm}p{1.7cm}p{1.85cm} p{1.7cm}}
			\toprule
			&   Data points (counts)&    Episode length (counts)&   Cumulative reward\newline (arb. units)&  Final \newline reward \newline (arb. units)\tabularnewline
			\hline\hline
			\emph{AE-DYNA} &500 &  3.28 $\pm$  1.26 & -1.44 $\pm$  1.10 &  0.04 $\pm$  0.06\\
			\emph{ME-TRPO} &450&  4.46 $\pm$  2.32 & -1.95 $\pm$  1.76 &  0.01 $\pm$  0.04\\
			\emph{NAF} &1074&  2.56 $\pm$  1.96 & -0.66 $\pm$  1.22 &0.00 $\pm$  0.02\\
			\emph{NAF2} &824& 2.64 $\pm$  1.65 & -0.57 $\pm$  0.92 &0.00 $\pm$  0.03\\
			%ae\_dyna & final &  0.04 &  0.06 \\
			%me\_trpo & final &  0.01 &  0.04 \\
			%naf\_double & final & -0.00 &  0.02 \\
			%naf\_single & final &  0.00 &  0.02 \\
			%ae\_dyna & nr &  4.28 &  1.26 \\
			%me\_trpo & nr &  5.46 &  2.32 \\
			%naf\_double & nr &  2.67 &  1.33 \\
			%naf\_single & nr &  2.56 &  1.26 \\
		\end{tabular}
		
	\end{table}
	\section{Conclusion}
	The presented reinforcement learning methods hold tremendous promise for automatizing set-ups typical in accelerators. One objective of this work was to provide some suggestions on making advanced deep reinforcement learning techniques work on a real set-up by adapting available methods. This was demonstrated on the FERMI FEL intensity optimization problem.\\
	Regularly, control problems in accelerators have short horizons, which makes \emph{DYNA}-style algorithms as the \emph{ME-TRPO} or our \emph{AE-DYNA} with their high representational power and excellent sample efficiency appealing choices.\\
	Complementary, the \emph{NAF2} algorithm presents a good alternative, revealing, as a model-free method, good asymptotic performance. As numerous cases in accelerator control can be captured assuming a quadratic dependence on the actions of the state-action value function, there is a broad spectrum of potential applications.\\
	To provide the possibility for other laboratories to profit from the stated methods, the code was released in \url{https://github.com/MathPhysSim/FERMI_RL_Paper}, and \cite{Hirlaender2020b}.
	
	\section*{Acknowledgement}	
	We want to thank Giulio Gaio, who made these experiments possible and helped set up and restore the FERMI FEL. Also, we are thankful for the valuable feedback from Alexander Scheinker on the draft of this publication. Finally, we want to thank Land Salzburg for their financial support to create the IDA Lab, which made this work possible.
	\appendix
	%	\appendices
%	\input{appendices.tex}
	
	\section{A Non-linear Standard Control Problem}
	To provide some transparency of these studies, we provide results on a famous classical standard control problem \cite{Furutaa}, the \emph{inverted pendulum}.\\
	It is a non-linear low dimensional unsolved continuous control problem. Unsolved means there is no threshold for the reward to terminate an episode. The episode length, the horizon, is set to 200 steps. The following several tests were carried out on the \emph{inverted pendulum} to demonstrate the improvements of the selected algorithms, mainly concerning the noise handling. It is of importance when dealing with measurements on real systems.\\
	For statistical significance, all shown results were obtained using five different seeds. The average value and the standard deviation (shaded) are plotted. For this study, we assume that the problem is successfully solved if the cumulative reward surpasses a threshold of -200. A dashed green line indicates the threshold in the corresponding figures.
	%	  and a rolling average of 15 steps is used for better visualisation.
	
	\subsection{NAF2 Details}\label{appendix:naf2}
\begin{figure}[!htbp]
		\centering
		\includegraphics*[width=0.475\textwidth]{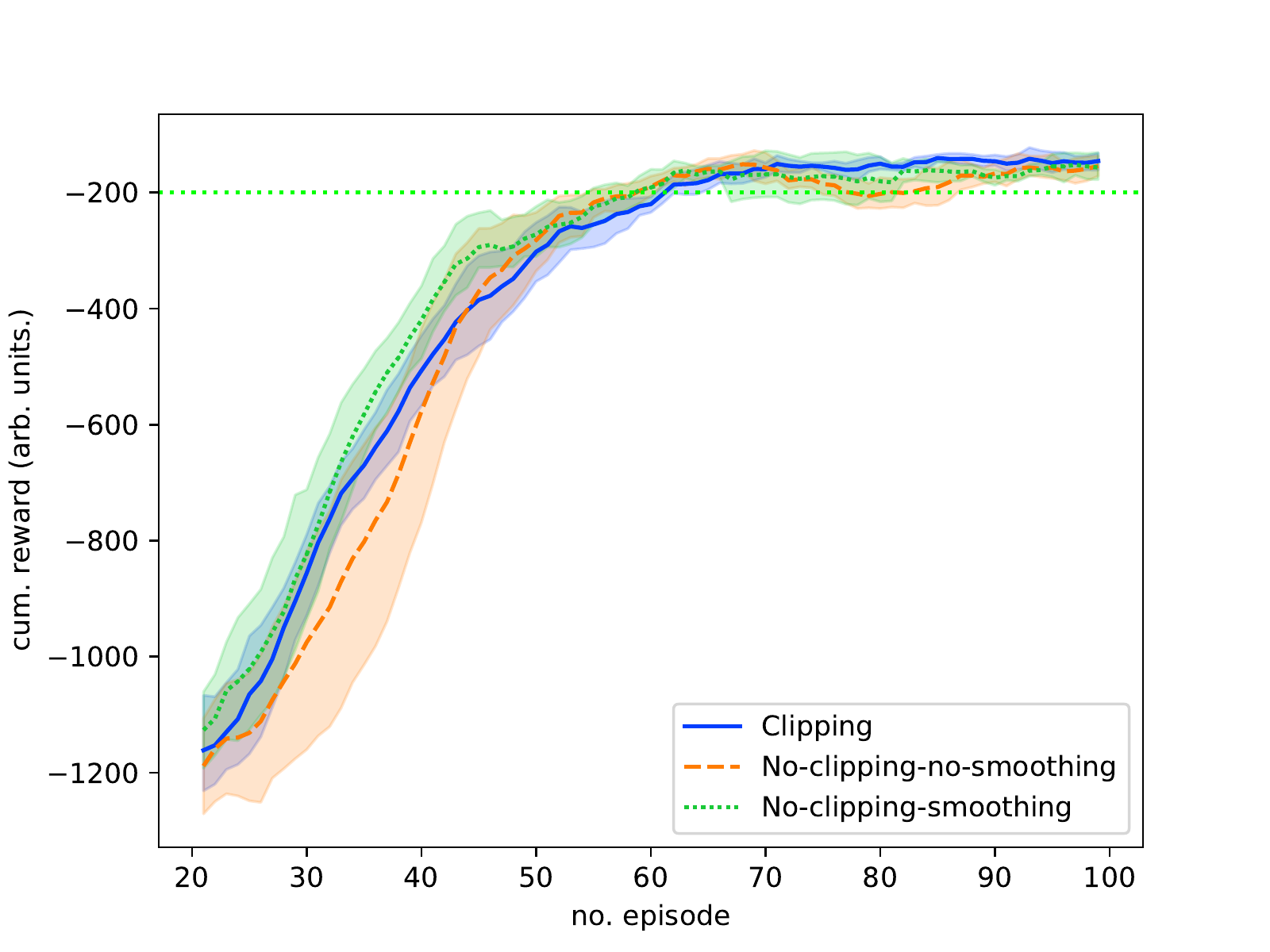}
		\caption{Cumulative reward of different \emph{NAF} implementations as discussed in the text on the \emph{inverted pendulum} without noise.}
		\label{fig:comparsion_smoothing_small}
	\end{figure}

	%We add artificial Gaussian noise $\gauss(0,0.05)$ on the state.
	We compare the different \emph{NAF} variants: \emph{Clipping}, \emph{No-clipping-smoothing}, \emph{No-clipping-no-smoothing}, where \emph{clipping} indicates the use of the double network, as introduced in \ref{ss:Normalized advantage function} and in all other cases a single network is used. 
	The term \emph{smoothing} indicates that a small clipped noise is added on the actions to stabilize the network training as:
	\begin{equation}
		\ba(\bs) = \text{clip}\left(\mu_{\theta_{\text{targ}}}(\bs) + \text{clip}(\epsilon,-c,c), \ba_{Low}, \ba_{High}\right),
	\end{equation}
	where $\epsilon \sim \mathcal{N}(0, \sigma)$. $c>0$ denotes the clipping coefficient and $ \ba_{Low}, \ba_{High}$ the minimal and maximal possible action. This method was used already in \cite{fujimoto2018addressing} to improve the deterministic policy gradient \cite{Silver2014}.
	The double network was used in \cite{fujimoto2018addressing,Haarnoja2018a} and is done in the following way:
	\begin{align}
		y(\reward_t,\bs_{t+1}, d_t) := \reward_t + \gamma (1 - d_t) \min_{i=1,2} V_{\theta_{i, \text{targ}}}(\bs_{t+1}),
	\end{align}
	with $d$ is 1 if the episode is finished and 0 otherwise.
	Then both are learned by regressing to this target (using the tuples from the data buffer $\mathcal D$):
	\begin{multline}
		L(\theta_i, {\mathcal D}) =\\ \mathbb E_{(\bs_{t},\ba_{t},r_t,\bs_{t+1},d_t) \sim {\mathcal D}}{
			\Bigg( Q_{\theta_i}(\bs_{t},\ba_{t}) - y(r_t,\bs_{t+1},d_t) \Bigg)^2
		},
		\\
		i \in \{1,2\}
	\end{multline}
	and the policy is obtained via $\max_\ba Q_{\theta_1}(\bs,\ba) = \mu_{\theta_1}(\bs)$.\\
	The results are shown in \ref{fig:comparsion_smoothing_small}. One sees the cumulative reward per episode for a training of 100 episodes. As mentioned, the curve labelled \emph{clipping} corresponds to the double network, including \emph{smoothing} and shows the best overall stability during the training yielding a high reward quickly. Also, the \emph{smoothed} single network, labelled \emph{No-clipping-smoothing}, shows good and comparable performance, except for the slightly decreased stability. The worst performance is achieved without smoothing and a single network (\emph{No-clipping-no-smoothing}), nevertheless the result is competing with state of the art model-free methods as \cite{BarthMaron2018} as the benchmark in the \emph{leaderboard} of openai gym \cite{Brockman2016}.
	%	\begin{figure}[!h]
		%		\centering
		%		\includegraphics*[width=0.5\textwidth]{Figures/Comparison_smoothing}
		%		\caption{Cumulative reward of different NAF implementations as discussed in the text.}
		%		\label{fig:comparsion_smoothing}
		%	\end{figure}
	
	%Note a subscript has been introduced, $j \in \{1 ... M\}$, with the view of an ensemble of $M$ ANNs, each with a distinct draw of ${\theta}_{anc}$. 

	% \tp{possible to just state equations you know, save space}
	\subsection{The Impact of Noise}\label{appendix:The impact of noise}
\begin{figure}[htbp]
		\centering
		\includegraphics*[width=0.475\textwidth]{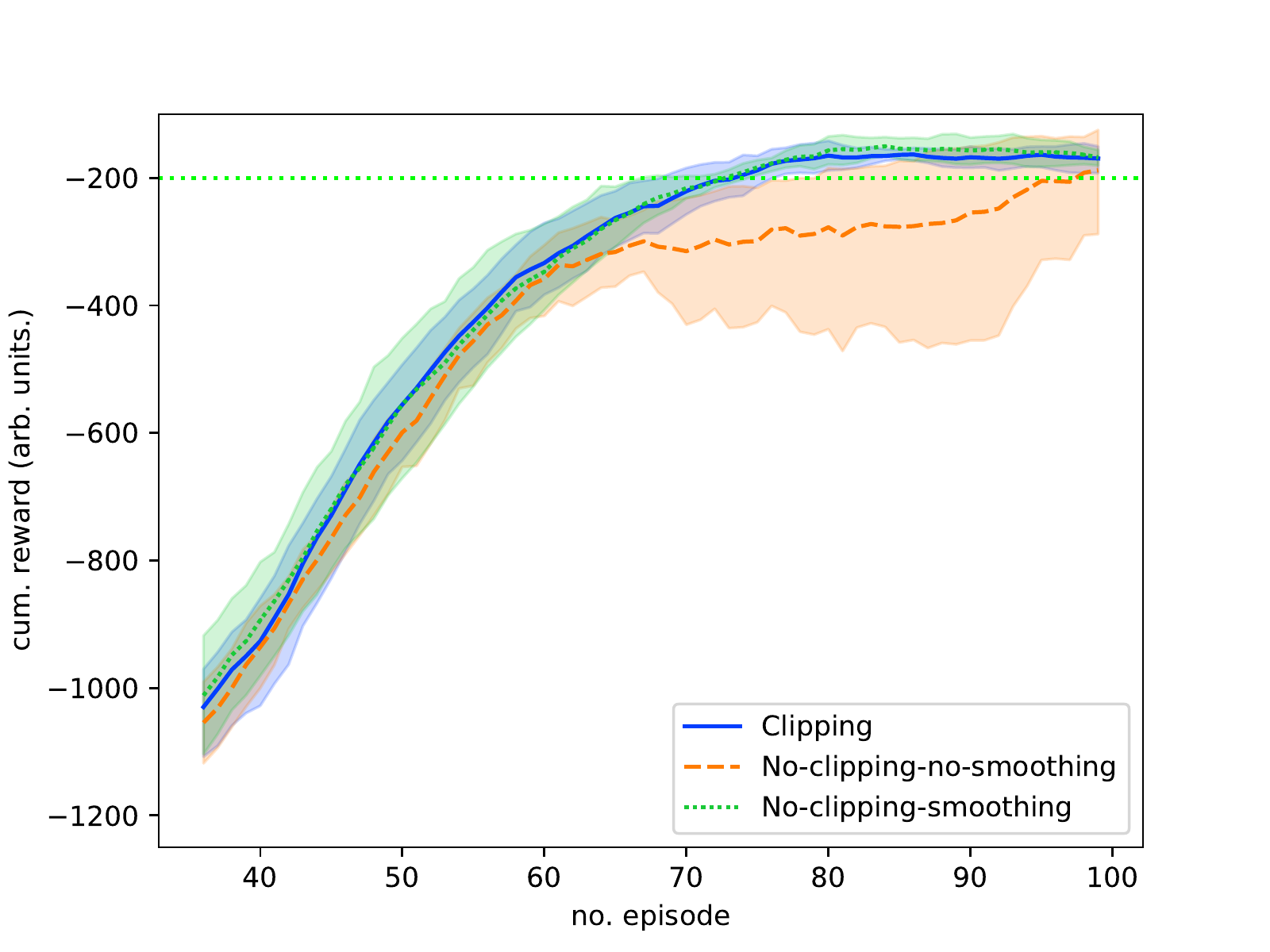}
		\caption{Cumulative reward of different \emph{NAF} implementations on the \emph{inverted pendulum} with artificial noise as discussed in the text.}
		\label{fig:comparsion_noise}
	\end{figure}

	A test adding large artificial Gaussian noise $\epsilon \sim \mathcal{N}(0, \sigma_\epsilon)$ with $\sigma_\epsilon=0.05$ in the normalized observation space on the states is presented in \ref{fig:comparsion_noise}. There the difference between the three methods becomes even more evident. The results are shown in \ref{fig:comparsion_noise}. After around 65 episodes, the single network without \emph{smoothing} (\emph{No-clipping-no-smoothing}) decreases before reaching the final performance at around 95 episodes, while  \emph{smoothing} prevents this performance drop in the other cases. 
	\subsubsection{Regression assuming homoscedastic Gaussian noise using `anchored ensembling'}
\begin{figure}[htbp]
		\centering
		\includegraphics*[width=0.475\textwidth]{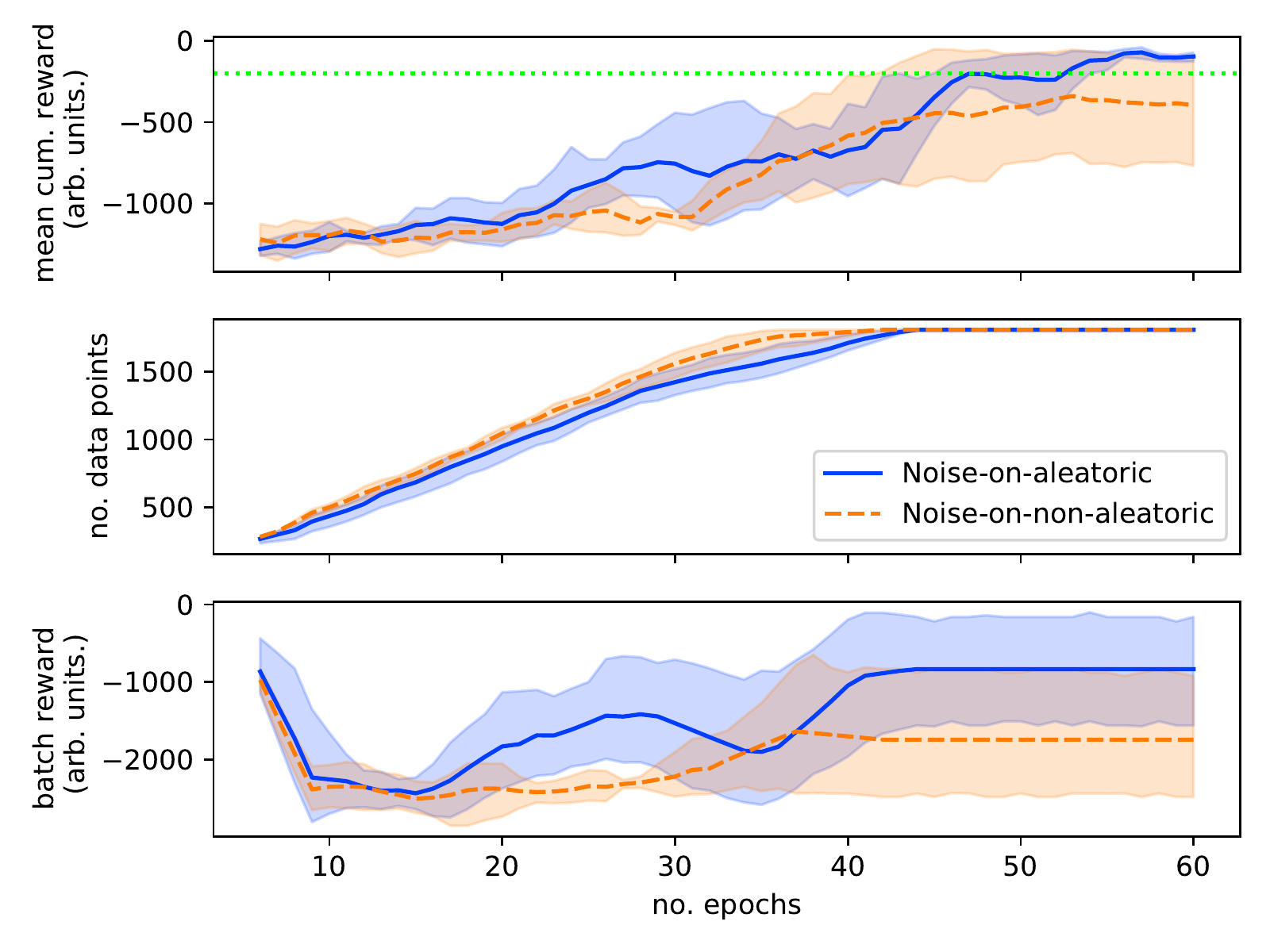}
		\caption{Cumulative reward of \emph{AE-DYNA-SAC} on the \emph{inverted pendulum} with artificial noise using the `anchor ensembling'.}
		\label{fig:comparsion_noise_ae_dyna}
	\end{figure}

\begin{figure}[htbp]
		\centering
		\includegraphics*[width=0.475\textwidth]{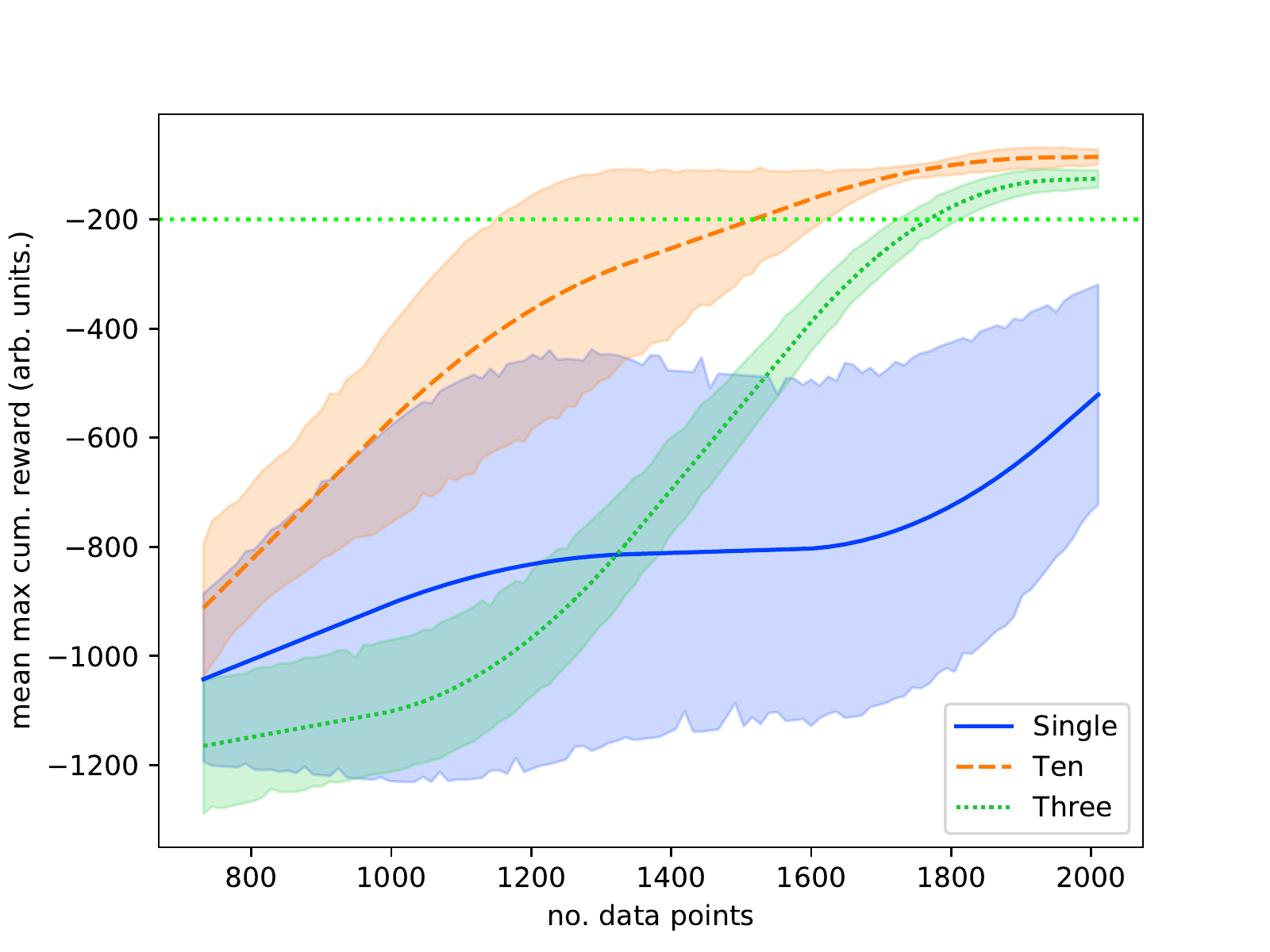}
		\caption{Varying number of models in the ensemble of the \emph{AE-DYNA-SAC} on the \emph{inverted pendulum}.}
		\label{fig:Compare_models_sizes}
	\end{figure}

	The Bayesian community and the RL community is spending more and more attention to
	new approaches to Bayesian inference. In the `anchored ensembling' technique a regularisation
	term is added to the loss function, which returns a point estimate of
	the Bayesian posterior the maximum a posteriori (MAP) parameter estimate. 
	Using an ensemble of networks and adding noise on either the targets of the regularisation term a distribution of MAP solutions  \cite{Gu2007, Chen2011, Bardsley2012, Pearce2018} is obtained, which mimics true posterior.\\
	Consider an artificial neural net containing parameters, ${\theta}$, making predictions for the dynamics and the reward \ref{eq:dynamics_model}, $\hat{f}_\theta$ ($f$ denotes the true unknown function), with $N$ data-points. If the prior is given by $P(\theta) = \mathcal{N}(\pmb{\mu}_{prior}, \pmb{\Sigma}_{prior})$, maximising the following returns MAP parameter estimates (details see \cite{Pearce2018}):
	\begin{equation}
		\label{eq_MAP_loglike_anc}
		{\theta}_{MAP}  = \text{argmax}_{{\theta}} \log( P_{\mathcal{D}}( \mathcal{D} | {\theta} ) ) - 
		\frac{1}{2} 
		\lVert \pmb{\Sigma}_{prior}^{-1/2} \cdot
		({\theta} - \pmb{\mu}_{prior}) \rVert^2_2.
	\end{equation}
	We replace $\pmb{\mu}_{prior}$ with some random variable $ {\theta}_{anc}$.
	%\begin{equation}
	%\label{eq_MAP_loglike_anc}
	%\pmb{f}_\text{MAP}(\pmb{\theta}_{anc})  = \text{argmax}_{\pmb{\theta}} \log( P_{\mathcal{D}}( \mathcal{D} | \pmb{\theta} ) ) - 
	% \frac{1}{2} 
	% \lVert \pmb{\Sigma}_{prior}^{-1/2} \cdot
	% (\pmb{\theta} - \pmb{\theta}_{anc}) \rVert^2_2
	%\end{equation}
	${\theta}_{anc} \sim \mathcal{N}(\pmb{\mu}_{prior},\pmb{\Sigma}_{prior} )$, to make it practical and
	assuming homoscedastic Gaussian noise of variance $\sigma^2_\epsilon$, we obtain a parametric form of the data likelihood. Taking $M$ models, where each model is indexed by$j \in \{1 ... M\}$, the MAP estimates are found by minimising ($\mathcal L_{j}$ denotes the loss of the $j^\text{th}$ model),
	\begin{equation}
		\label{eqn_anch_loss_matrix}
		\mathcal L_{j} =  
		\frac{1}{N} \lvert \lvert f - \hat{f}_{\theta, j} \rvert \rvert ^2_2
		+ \frac{1}{N} \lvert \lvert \pmb{\Gamma}^{1/2} \cdot (\theta_j - \theta_{anc,j}) \rvert \rvert ^2_2.
	\end{equation}
	$\pmb{\Gamma}$ is diagonal regularisation matrix. The $i^{th}$ diagonal element is the ratio of data noise of the target variable to prior variance for parameter $\theta_i$:
	\begin{equation}
		\label{eqn_anch_loss_init}
		\text{diag}(\pmb{\Gamma})_i = \frac{\sigma^2_\epsilon}{\sigma^2_{prior_i}}.
	\end{equation}
	Using the anchors and $\sigma_\epsilon=0.05$ in the dynamics model stabilizes the training of the \emph{AE-DYNA} as illustrated in \ref{fig:comparsion_noise_ae_dyna}. The mean cumulative reward during the training on 10 test episodes on the real environment is shown in the upper plot. It should indicate the result if the training is stopped at this training epoch. One cannot observe this quantity during real training unless one does costly performance measurements while training. Respecting the aleatoric (\emph{Noise-on-aleatoric}) helps to reach the target much quicker, exhibiting less variation compared to the standard use of an ensemble (\emph{Noise-on-non-aleatoric}). An epoch consists of 3000 iterations of the \emph{SAC}.\\
	The lower plot of \ref{fig:comparsion_noise_ae_dyna} shows the batch rewards a measured during the data collection, which is observable during the training.
	\ref{fig:Compare_models_sizes} shows the impact of the number of models onto the performance on the \emph{inverted pendulum}. The maximum cumulative reward averaged over five different runs tested on the real environment during the training is visualized in dependence of the number of data-points. A number of three models (label \emph{Three}) shows a good trade-off between performance and training time. A single network might not converge or too slowly (label \emph{Single}) and a model of ten shown fast and stable performance in this case (label \emph{Ten}).
	\subsection{NAF versus AE-DYNA}
	\begin{figure}[htbp]
		\centering
		\includegraphics*[width=0.475\textwidth]{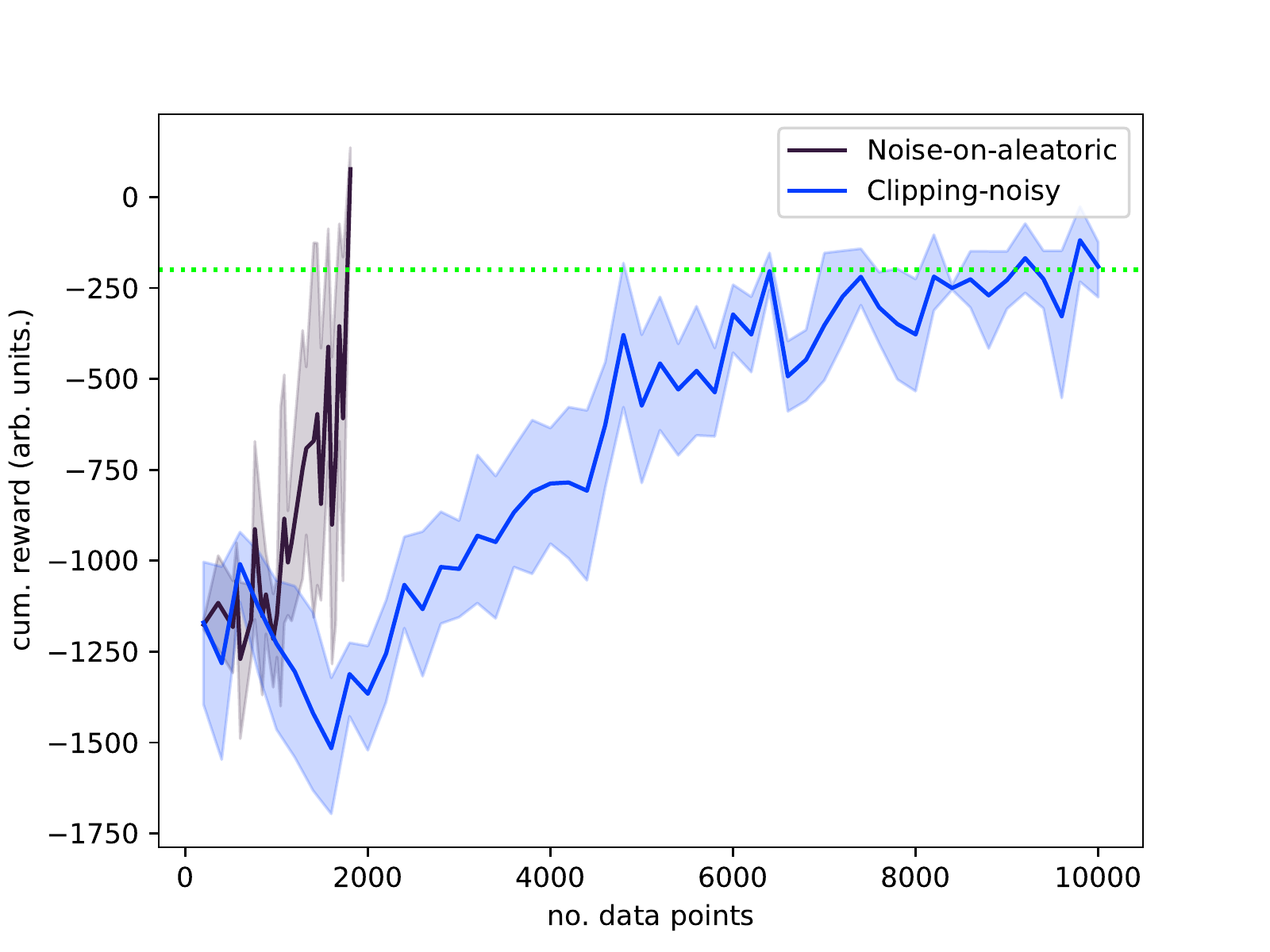}
		\caption{The comparison of the \emph{NAF2} and the \emph{AE-DYNA-SAC} on the noisy \emph{inverted pendulum}.}
		\label{fig:comparsion_NAF_AE-DYNA}
	\end{figure}
	Finally, \ref{fig:comparsion_NAF_AE-DYNA} demonstrates the sample-efficiency of the \emph{AE-DYNA-SAC} and the \emph{NAF} algorithm on the noisy \emph{inverted pendulum}. \emph{AE-DYNA-SAC} converges below 2000 data-points (without noise even below 800), and the \emph{NAF2} starts to perform equally 10000 data-points surpassing the -200 reward threshold. One clearly sees the increased sample-efficiency on this problem using the \emph{AE-DYNA} in contrast to the \emph{NAF2}.

	\bibliography{Bibliography}
\end{document}